\documentclass[10pt]{article}
\ifdefined\JubileeAnonymous
  \usepackage{tmlr}
\else
  \usepackage[preprint]{tmlr}
\fi
\usepackage{amsmath,amssymb,booktabs,graphicx}
\usepackage{hyperref,url,adjustbox}
\usepackage[section]{placeins}
\AddToHook{cmd/subsection/before}{\FloatBarrier}
\AddToHook{cmd/subsubsection/before}{\FloatBarrier}
\hypersetup{hidelinks}
\setcitestyle{aysep={},citesep={,},yysep={,}}
\usepackage[inline]{enumitem}
\setlist{nosep}
\graphicspath{{figures/}}
\makeatletter
\newcommand{\unmarkedfootnote}[1]{\begingroup
  \renewcommand\@makefntext[1]{\noindent##1}\footnotetext{#1}\endgroup}
\makeatother
\title{JUBILEE: A Benchmark Environment for\\Off-Policy Evaluation in Debt Collections}
\ifdefined\JubileeAnonymous
  \author{Anonymous authors}
  \hypersetup{pdfauthor={Anonymous authors},pdftitle={JUBILEE}}
\else
  \author{\name Diego Silva Prado \email diego@alphasynthesis.com \\
\addr Alpha Synthesis Labs
\AND
\name Tiago Amaral Penteado de Jesus \email tiago@alphasynthesis.com \\
\addr Alpha Synthesis Labs
\AND
\name Bruno Caldas de Souza \email caldas@alphasynthesis.com \\
\addr Alpha Synthesis Labs
\AND
\name Cristiano Bevitori Maffia De Oliveira \email cbmo@alphasynthesis.com \\
\addr Alpha Synthesis Labs}

  \hypersetup{pdfauthor={Diego Silva Prado and coauthors},pdftitle={JUBILEE}}
\fi
\begin{document}
\maketitle
\ifdefined\JubileeAnonymous\else
\unmarkedfootnote{The code is available at
\url{https://github.com/alpha-synthesis/jubilee}\newline and the data at
\url{https://doi.org/10.5281/zenodo.24008544}.}
\fi
\begin{abstract}
Off-policy evaluation (OPE) estimates the value a candidate decision policy would achieve, using only data logged by the policy already in production. In debt collections, unrecorded factors can influence both actions and outcomes. This paper presents JUBILEE, a benchmark environment built around a debt-collections simulator that generates million-loan panels with competing risks, rare recovery events, and hidden factors driving both outcomes and assignment. The evaluation includes an 80-run comparison of tabular and gradient-boosted estimators under two synthetic outcome mechanisms and a 190-run parameter grid. Independent policy simulations provide reference values with estimated Monte Carlo uncertainty for a synthetic reward score and for simulated cash. With hidden factors influencing action assignment, several tabular methods agree closely while underestimating the field-visit policy's score, an error that also appears with boosted models. Intervals from tabular doubly robust estimation usually contain the reference without hidden assignment dependence but mostly miss it when that dependence is present. Estimates from randomized decisions do not show the same large error, although their limited precision leaves policy choice uncertain. The decision consequences depend on how policy value is defined. The field-visit policy ranks last under the score and first under simulated cash, which excludes channel costs. Under cash, log-based estimators miss this best candidate in nearly every run with hidden assignment dependence in the paired study. JUBILEE supports reproducible study of these failure modes under specified collections mechanisms.
\end{abstract}
\section{Introduction}\label{sec:intro}

Debt collections is becoming a reinforcement-learning domain. Discount
policies are learned and deployed at a financial
institution~\citep{collectionsrl2025}.
Off-policy evaluation (OPE) estimates a candidate policy's value from
logs generated by an incumbent. Proprietary collections records can be
difficult to share~\citep{kriebel2020}, while public credit data do not
supply the counterfactual contact-and-offer outcomes needed to measure
OPE error directly.

Even detailed logs can omit hardship, contactability or other information
used to assign actions. The resulting confounding can make observational
associations differ from intervention effects. Simulation supplies
references under specified mechanisms, with precedents in healthcare and
adapted MDPs~\citep{namkoong2020offpolicy,brunssmith2021causality,
kausik2024offline}. Collections adds contact channels, settlement offers,
rare recoveries and competing terminal events to that evaluation setting.

JUBILEE, the benchmark environment introduced in this paper, turns those
failure conditions into experimental parameters.
The contribution is an environment and protocol, evaluated through three
complementary analyses:

\begin{itemize}
\item Independent reference populations quantify simulation uncertainty,
including paired policy differences, so estimator error is read against
reference precision.
\item Under hidden coupling, tabular and gradient-boosted log-based
estimators underestimate the field-visit policy by about one fifth of its
score. Tabular DR intervals usually contain the reference in controls
but mostly miss it under coupling. Agreement among tabular estimators is
algebraic, while boosted estimators disagree in controls and coupled
conditions alike. Neither pattern alone identifies the error.
\item Estimators restricted to flagged randomized decisions avoid this
error at about three times the interval width, and refitting the direct
method (DM) exposes omitted fitting uncertainty. The underestimated policy
is the worst candidate under the score and the best under simulated cash,
gross of channel costs. Under cash, coupling therefore leads tabular DR
to miss the best candidate in nearly every run of the paired study.
\end{itemize}

The parameters are deliberately not calibrated to any portfolio, and every
claim concerns estimator behavior under known conditions. JUBILEE's 4\%
recovery target controls event rarity. Results use the score defined in
Section~\ref{sec:scoredefinition}, and Section~\ref{sec:scorecash} repeats
them under simulated cash. Shared seeds and fixed candidate rankings
further bound generalization.

This paper is organized as follows. Section~\ref{sec:related} reviews
related work. Section~\ref{sec:testbed} presents the benchmark environment,
and Section~\ref{sec:protocol} describes the experimental protocol.
Section~\ref{sec:results} reports the results, while
Section~\ref{sec:discussion} discusses their implications and limitations.
Section~\ref{sec:conclusion} concludes. The appendices provide configuration
details, complete results and validation checks.

\section{Related Work}\label{sec:related}

The estimators graded here are the standard OPE toolkit. They include
inverse-propensity weighting from Horvitz--Thompson
theory~\citep{horvitz1952generalization}, its self-normalized
variant~\citep{swaminathan2015self}, and doubly robust (DR)
estimation~\citep{dudik2011doubly}, with cross-fitted
nuisances~\citep{chernozhukov2018double}. JUBILEE's contribution therefore
concerns the environment and evaluation protocol.

Existing benchmarks provide infrastructure for these comparisons.
COBS compares OPE estimators across RL settings~\citep{voloshin2021cobs},
while DOPE supplies policy-dataset benchmarks~\citep{fu2021dope}.
Open Bandit complements these resources with randomized recommendation
logs~\citep{saito2021obp}, and SCOPE-RL supplies evaluation
software~\citep{kiyohara2023scoperl}. Alongside this benchmarking work,
controlled confounding has direct precedents in sequential sensitivity
bounds in healthcare~\citep{namkoong2020offpolicy}, model-based and
model-free bounds~\citep{brunssmith2021causality}, identification under
different temporal assumptions~\citep{kausik2024offline}, structural
counterfactual models~\citep{oberst2019counterfactual}, and sensitivity-bound
estimation for robust policy learning~\citep{hess2026efficient}.
JUBILEE's generated-confounding parameter differs from an estimator's
sensitivity parameter.

A complementary line extends evaluation information through
counterfactual annotations.
\citet{tang2023isplus} augment IS, and CANDOR~\citep{mandyam2024candor}
develops DR variants. Following this line, JUBILEE includes
annotation-augmented estimation in its panel, drawing annotations from
the privileged simulator rather than from people. It also examines when
the tabular estimator construction cancels their contribution.

These methodological connections also extend to finance. Simulators such
as ABIDES~\citep{byrd2020abides} and synthetic-data
research~\citep{assefa2020generating} motivate reusable environments.
Within collections, prior work includes proprietary-data
studies~\citep{kriebel2020,collectionsrl2025},
public RL code~\citep{ycenycollections} and LendingClub-based policy
simulation with recovery, costs and net profit~\citep{sanzmartin2026adaptive}.
Against this background, JUBILEE grades OPE from logged actions while varying hidden
assignment, noise, exploration and annotation quality.
Table~\ref{tab:comparison} brings these methodological and domain
antecedents together. Different objectives
and temporal assumptions preclude benchmark equivalence, and the combination
of simulation, controlled confounding and references carries no general
priority claim.

\begin{table}[!htbp]
\centering
\caption{Related evaluation settings and their scope.}
\label{tab:comparison}
\footnotesize
\begin{tabular}{p{0.18\textwidth}p{0.15\textwidth}p{0.26\textwidth}p{0.29\textwidth}}
\toprule
work & domain and estimand & confounding or randomization & reference and evaluation role \\
\midrule
\citet{namkoong2020offpolicy} & healthcare simulations, sequential value bounds & unobserved sequential confounding and sensitivity assumptions & simulated policy values for evaluating bounds, released experimental code \\
\citet{brunssmith2021causality} & adapted MDPs, finite-horizon value bounds & generated latent confounders, temporal persistence matters & simulated values for comparing model-based and model-free bounds \\
\citet{kausik2024offline} & gridworld and sepsis, evaluation and improvement & alternative temporal confounding structures & simulated values and returns for evaluating confounding-aware methods, released code \\
\citet{hess2026efficient} & synthetic and semisynthetic policy learning & latent confounding plus sensitivity bounds, random policy comparator & known synthetic potential outcomes and regret, random comparator is not a logged pilot \\
\citet{sanzmartin2026adaptive} & collections, learned-policy net profit & risk-based transition model and action-cost variation & simulated rollouts for comparing policies \\
JUBILEE & collections, initial action followed by behavior policy & latent borrower and regional factors, flagged randomized logging subset & independent Monte Carlo policy references, tabular and boosted regimes, synthetic score and simulated cash \\
\bottomrule
\end{tabular}
\end{table}

\section{The JUBILEE Benchmark Environment}\label{sec:testbed}

\begin{figure}[!htbp]
  \centering
  \includegraphics[width=\textwidth]{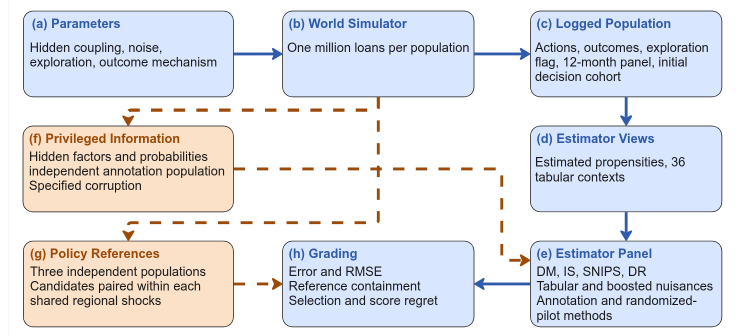}
  \caption{The JUBILEE evaluation protocol.}
  \label{fig:pipeline}
\end{figure}

JUBILEE wraps a synthetic debt-collections simulator in an evaluation
protocol. Separate pathways supply logged observations, privileged
annotations and policy references. The paired-study configuration is
shown in Figure~\ref{fig:pipeline}. Estimators receive observable logs,
while independent simulations supply grading values.
The equations in this section define JUBILEE's synthetic model and score.

\subsection{The simulated world}\label{sec:scoredefinition}

Each world contains one million loans. Each loan moves through a monthly
state machine, from current through delinquent to late-stage delinquency,
with two competing terminal events, recovery and charge-off. Both studies log 12 months for a nine-month reward window.
Under the released simulator's full-settlement rule, an accepted offer is
paid in full, the discounted remainder is written off and the loan closes
as cured. Actions comprise no contact, SMS, app message, phone or field
visit, with offers in $[0,0.55]$.
The benchmark's decision score is written as:

\begin{equation}
  r_i \;=\; \mathbf{1}\!\left[\text{recovered in months } m..m{+}8\right]
  \cdot \text{balance}_i \cdot (1 - d_i)
  \label{eq:reward}
\end{equation}

where $r_i$ is the score for loan $i$, $m$ is the decision month,
$\text{balance}_i$ is the balance at that decision and $d_i$ is the
offered discount fraction. The indicator $\mathbf{1}[\cdot]$ records recovery
within the nine-month window from $m$ through $m+8$. The score applies
that discount to any first recovery in the window, including recovery
under a later offer. Continuation after the decision month is common to
all candidates, so the score separates candidates largely through their
initial discounts. Section~\ref{sec:scorecash} repeats the evaluation with
simulated cash, the payments of months $m$ to $m+8$, which under full
settlement is a single payment of the balance at recovery net of the
discount offered in that month. Values, errors and regrets use this
US\$-per-decision score unless labeled as cash. It excludes channel costs and borrower
welfare and differs from payments received. The horizon and offer bound are experimental settings.

Recovery follows a logistic model whose monthly probability for decision
$i$ is given by:

\begin{equation}
  \begin{aligned}
  \operatorname{logit} p_i ={}& \operatorname{logit}(\tau) + \gamma_{a_i}(\mathrm{dpd}_i)
    + \beta(1-e^{-\kappa d_i}) + f(z_i) \\
    &+ 0.55\,u_i^{\mathrm{res}} - 0.60\,u_i^{\mathrm{fric}}
     - 0.45\,m_{r(i)} \\
    &+ 0.35\,\mathbf{1}[a_i \ne \mathrm{none}]\,u_i^{\mathrm{cont}}
     + \sigma\,\xi_i
  \end{aligned}
  \label{eq:hazard}
\end{equation}

where $p_i$ is monthly recovery probability and $\operatorname{logit}$
the log-odds transform. The baseline monthly hazard is $\tau$, $a_i$ is the
contact action, $d_i$ its discount and $\mathrm{dpd}_i$ days past due.
The log-odds channel effect is $\gamma_{a_i}(\mathrm{dpd}_i)$, while $\beta$ and
$\kappa$ control the magnitude and saturation of the discount response.
The observed covariates $z_i$ enter through $f(z_i)$, specified in
Appendix~\ref{sec:appendix}.

The components $u_i^{\mathrm{res}}$, $u_i^{\mathrm{fric}}$ and
$u_i^{\mathrm{cont}}$ represent resilience, payment friction and
contactability. The mapping $r(i)$ selects the loan's region and
$m_{r(i)}$ its shared shock. The term $\sigma\xi_i$ scales a standard
Gaussian log-odds shock $\xi_i$ by $\sigma$. The indicator activates
contactability only when $a_i$ differs from $\mathrm{none}$, the no-contact
action. The latent factors are distributed as:

\begin{equation}
  u_i \sim \mathcal{N}\!\left(0,\, \alpha^2 I_3\right),
  \qquad
  m_r \sim \mathcal{N}\!\left(0,\, (0.45\,\alpha)^2\right)
  \label{eq:hidden}
\end{equation}

where $\mathcal{N}$ is a Gaussian distribution with mean zero, and
$I_3$ is the $3\times3$ identity matrix. The vector $u_i$ collects the
three independent borrower factors, each with standard deviation $\alpha$.
The shock $m_r$ for region $r$ has standard deviation $0.45\alpha$. These
distributions and coefficients are synthetic assumptions, without
portfolio calibration.

A coupling switch controls whether hidden factors influence assignment.
With coupling enabled, harder cases receive more contact. Disabling it
retains hidden outcome variance but removes that assignment dependence,
providing a negative control at the experiment's resolution. Experimental
settings are specified in Section~\ref{sec:protocol}.

\subsection{Logging policy and exploration floor}\label{sec:floor}

JUBILEE defines the baseline incumbent $\pi_b$ through its channel and
discount-bin distributions. Its nominal probabilities on the action grid, before offer rounding,
are written as:

\begin{equation}
 \pi_b(a\mid\mathrm{dpd}_i,s_i)=
 \begin{cases}
 p_b(c\mid s_i)\,q(k\mid\mathrm{dpd}_i),
    & a=(c,k),\ c\ne\mathrm{none},\\
 p_b(\mathrm{none}\mid s_i), & a=\mathrm{none}
 \end{cases}
 \label{eq:incumbent}
\end{equation}

where $a$ is the action for loan $i$, $c$ its channel and $k$ its discount
bin. The functions $p_b$ and $q$ give the channel and discount-bin
probabilities, respectively. Days past due are $\mathrm{dpd}_i$, and
$s_i$ is dimensionless assignment intensity. The no-contact action has zero discount.
With hidden coupling enabled, the benchmark's intensity is given by:

\begin{equation}
  s_i \;=\; \operatorname{clip}\!\big(\mathrm{dpd}_i/180
  \;+\; u_i^{\mathrm{fric}} \;-\; 0.4\, u_i^{\mathrm{cont}},\; [-2, 2]\big)
  \label{eq:intensity}
\end{equation}

where $u_i^{\mathrm{fric}}$ is the hidden friction factor and
$u_i^{\mathrm{cont}}$ is the hidden contactability factor. The operator $\operatorname{clip}$
truncates intensity to $[-2,2]$. Higher friction or delinquency increases
$s_i$, which changes the channel probabilities through the
normalized exponential weights~\citep{mcfadden1974conditional}, written as:

\begin{equation}
 p_b(c\mid s_i)=
 \frac{w_c\exp(\eta_c s_i)}
 {\sum_{j\in\mathcal C}w_j\exp(\eta_j s_i)}
 \label{eq:channelpolicy}
\end{equation}

where $w_c$ is the base weight and $\eta_c$ its intensity coefficient.
The set $\mathcal C$ contains the five channels, including no contact,
and $j$ indexes the denominator's sum over channels.
Table~\ref{tab:coefficients} gives their weights and coefficients. With
coupling off, all $\eta_c$ are zero. The discount component is given by:

\begin{equation}
 q(k\mid\mathrm{dpd}_i)=\Pr(D_i\in I_k\mid\mathrm{dpd}_i)
 \label{eq:discountbins}
\end{equation}

where $D_i$ is the offer before rounding and $I_k$ its $k$-th discount
interval. Bins are five percentage points wide, with the last including
the 55\% cap. The offer combines a scaled Beta draw with an adjustment
for delinquency. Appendix~\ref{sec:appendix} specifies this law, expands
$q$ and describes rounding.

Flagged randomization adds an exploration component with known assignment
probabilities~\citep{horvitz1952generalization,li2011unbiased}. On
explorable states, the resulting logging policy is written as:

\begin{equation}
  \tilde{\pi}_b(a \mid \mathrm{dpd}_i,s_i)
  \;=\; (1 - \varepsilon)\, \pi_b(a \mid \mathrm{dpd}_i,s_i)
  \;+\; \frac{\varepsilon}{44}\, \mathbf{1}[a \text{ active}]
  \label{eq:floor}
\end{equation}

where $\tilde\pi_b$ is the logging policy with exploration, $\pi_b$ is
the baseline incumbent and $\varepsilon$ the randomized share. The indicator selects active
contact actions. Exploration draws a channel uniformly from the four
contact channels and an independent discount uniformly on $[0,0.55)$.
The eleven equal-width bins therefore give each of the 44 channel--offer
cells nominal randomization probability $\varepsilon/44$. Explorable
states comprise delinquent loans and late-stage delinquency within the
eligibility limit of Section~\ref{sec:protocol}. Outside these states,
$\tilde\pi_b=\pi_b$. At $\varepsilon=0$, pilots are undefined
because no flagged randomized rows exist. This construction links hidden
factors to assignment when coupling is enabled and supplies flagged
randomized decisions for evaluation.

\subsection{Reference values by policy simulation}\label{sec:intervention}

JUBILEE simulates each candidate policy directly to obtain reference
values for assessing OPE estimates. The candidate $\pi$ acts at month
zero, and subsequent decisions follow the same logging policy
$\tilde\pi_b$ through the nine-month window. The corresponding
policy-value expectation~\citep{dudik2011doubly} is written as:

\begin{equation}
 V(\pi\mid M)=\mathbb{E}_{\pi,\tilde\pi_b}\!\left[
  \mathbf{1}\{\text{recovered in months }0\!:\!8\}
  B_0(1-d_0)\mid \text{initially explorable},M
 \right]\label{eq:estimand}
\end{equation}

where $V(\pi\mid M)$ is the expected score per initially explorable loan,
conditional on the regional-shock vector $M$. The quantities $B_0$ and
$d_0$ are the month-zero balance and discount, and the indicator selects
first recovery in months 0--8. The expectation averages borrower and
outcome variation. The reference estimates this population expectation
with Monte Carlo error, rather than an exact counterfactual for the
logged finite portfolio. For cash, payments in months 0--8 replace the
score inside the expectation.

Both studies use independent populations for logs, annotations and
reference values, conditional on common regional shocks. Within each
reference replicate, candidates share borrowers, latent factors and
random draws to improve the precision of their value differences. For a
fixed configuration, seed, replicate and batch size, only actions and
the resulting trajectories differ. Changing the batch size changes the
population.

\subsection{The estimator view and the data gate}

The tabular view has 36 contexts, crossing capped 30-day delinquency
buckets with logged-balance terciles, and 45 actions, comprising the null
action and the active offer bins. Balance cut points are learned from
logs and reused across reference and annotation populations. Tabular estimators use empirical within-context action frequencies
computed from the logged cohort. This coarse
view can leave observed assignment variation unadjusted as well as
omitting hidden factors. Boosted models use richer observed covariates
and cross-fitted propensity classifiers (Section~\ref{sec:protocol}).

A hard data gate separates what estimators may read from what grading
may read. It excludes hidden factors, true recovery probabilities and the
incumbent rule from estimator inputs. Privileged information supports
only grading and the specified annotation procedure, keeping both
separate from observational policy evaluation.

\section{Experimental Protocol}\label{sec:protocol}

The 190-run grid maps sensitivity within a tabular construction. The
80-run paired study compares tabular and boosted estimators under two
outcome mechanisms, using independent references to assess precision,
intervals and selection. Both evaluate the same initial-action score and
simulated cash.

\subsection{Parameter grid}\label{sec:archiveprotocol}

The grid evaluates sensitivity to exploration, outcome noise and confounding
through seventeen selected parameter configurations, repeated over simulation
seeds for a total of 190 runs. These configurations combine confounding strength
$\alpha\in\{0.2,0.4,0.6,1.0,1.4,1.8\}$, outcome noise
$\sigma\in\{0.15,0.25,0.40,0.60\}$ and exploration
$\varepsilon\in\{0,0.08\}$. Along the confounding axis, $\sigma=0.25$
is held fixed, while the noise axis holds $\alpha=0.6$. Two additional
configurations combine higher confounding and noise, and a coupling-off
control retains hidden variance at $\alpha=0.6$. Together, these comparisons
form the grid reported in Table~\ref{tab:gridvalues}.

To keep recovery events rare across the grid, each cell's baseline hazard
was calibrated on 50,000-borrower development worlds to an internal annual
recovery target of $4.0\%\pm0.3$ percentage points. Post-charge-off recovery
rates and debt-purchase prices concern different events and
populations~\citep{cfpb2021creditcard,ftc2013debtbuying}. Cells already within
the calibration tolerance retained the strength-0.6 hazard of 0.001700,
which five cells therefore share in Table~\ref{tab:gridvalues}. The stronger
confounding and noise cells were retuned, while the coupling-off control's
hazard is 11\% above its coupled counterpart. Under the released simulator,
these hazards yield cell-mean annual recovery of 3.9 to 4.4\%, with one cell
above the calibration tolerance.

This recalibration affects the interpretation of comparisons across cells.
Each retuned cell jointly specifies the varied parameter and its
compensating baseline hazard. Consequently, exploration pairs isolate a
fixed-hazard exploration effect only where both members share a hazard.
Section~\ref{sec:explorationscore} examines that effect separately.

Within each configuration, the target $\pi_{\mathrm{visit}}$ assigns a visit
at discount 0.275, providing a thin-support challenger, while
$\pi_{\mathrm{sms}}$ assigns SMS at 0.125, closer to incumbent behavior.
Both targets follow the intervention scheme of
Section~\ref{sec:intervention} on the initial explorable cohort of
approximately 744{,}000 loans. Their offers sit at bin centers, while
matching logged offers vary within bins.

Each run evaluates both targets for one configuration and simulation seed. The seven
confounding-axis cells with exploration use 20 seeds each, while the
remaining cells use 5. This allocation gives greater precision to the
confounding comparison. With 20 seeds, the pilots' seed-level dispersions
of 46--86~US\$ imply approximate 80\%-power detection thresholds of
29--54~US\$ at the 5\% level, conditional on those dispersions. Seeds are
shared across cells, preserving regional-shock dependence without
guaranteeing identical portfolios. To assess estimation error within each
run, three independent million-borrower reference populations provide the
comparison values in Table~\ref{tab:gridvalues}, following the reference
design detailed in Section~\ref{sec:campaign}.

\subsection{Paired design and reference precision}\label{sec:campaign}

The paired design compares experimental conditions using shared simulated
populations and random draws, so differences can be assessed within the same
seed block. Each block contains worlds A and B with assignment coupling off
and on. Within each condition, tabular and gradient-boosted estimators use
the same logged data and policy references. The same simulated trajectories
are also evaluated under both the synthetic score and simulated cash.
Independent reference populations provide policy values and quantify their
Monte Carlo uncertainty, with candidate policies paired within each population.

\subsubsection{Worlds, candidates and shared seed blocks}

The paired study crosses worlds A and B with the assignment coupling
defined in Section~\ref{sec:scoredefinition} switched off and on, using
$\alpha=1$, $\sigma=0.25$ and $\varepsilon=0.08$. When coupling is on,
intensity $s_i$ affects the channel probabilities, whereas with coupling
off, the coefficients $\eta_c$ are set to zero and channels follow the
base weights $w_c$. This switch controls
whether hidden factors influence channel assignment, while retaining
their effects on recovery in both conditions.

In world A, channel effects
are fixed across delinquency stages, and the baseline hazard is the
grid's strength-1.0 value, held fixed across coupling conditions.
By contrast, world B has stage-dependent active-channel effects that
change at 90 days past due. Its baseline hazard is calibrated in
development to match A's annual recovery rate and is also held fixed
across coupling conditions within B. Table~\ref{tab:campaignworlds}
reports the log-odds channel effects and monthly baseline hazards.
These are synthetic mechanism contrasts, without estimated real-world
channel efficacy.

\begin{table}[!htbp]
\centering
\caption{Recovery mechanisms and baseline hazards.}
\label{tab:campaignworlds}
\small
\begin{tabular}{lrrr}
\toprule
channel & A, all stages & B, before 90 days & B, from 90 days \\
\midrule
none & $-0.85$ & $-0.85$ & $-0.85$ \\
SMS & $+0.10$ & $+0.30$ & $-0.10$ \\
app message & $+0.18$ & $+0.30$ & $0.00$ \\
phone & $+0.35$ & $+0.30$ & $+0.45$ \\
field visit & $+0.48$ & $+0.30$ & $+0.65$ \\
baseline hazard $\tau$ & 0.001421 & \multicolumn{2}{c}{0.001426} \\
\bottomrule
\end{tabular}
\end{table}

Because the coupling switch also changes observed delinquency tilt,
action support and incumbent continuation, on--off contrasts do not
isolate a pure change in hidden information. For these comparisons,
twenty seeds form paired blocks across all four conditions, sharing
populations, latent factors, regional shocks and random draws.
Within each coupling condition, worlds A/B also share initial logged
actions. The 80 runs comprise 20 independent blocks, with disjoint
development seeds.

Policy selection in these blocks compares the incumbent $\pi_b$, the two
constant targets and $\pi_{\mathrm{stage}}$, which assigns SMS at discount
0.025 before 90 days past due and a visit at 0.275 thereafter.
All challengers use the intervention scheme in
Section~\ref{sec:intervention}, and eligibility requires initial
delinquency or late-stage delinquency of at most 180 days.
The stage boundary aligns with the tabular context. These candidates and
eligibility rules are experimental settings, without operational
optimality claims.

\subsubsection{Reference precision and analysis}

For these comparisons, each run uses three independent
million-borrower reference populations, conditional on common regional
shocks, with candidates paired within populations. A development pilot
selects the smallest replicate count meeting per-run Monte Carlo
standard-error (SE) targets of 2.0~US\$ for values and 1.0~US\$ for paired
gaps within the compute ceiling. For policy comparisons, a
best-versus-runner-up gap exceeding 1.96 paired reference SEs is labeled
resolved as a pointwise precision diagnostic, without simultaneous winner
guarantees. Variation across seed blocks is reported separately from
this Monte Carlo uncertainty.

Using these references, errors are defined as estimates minus references.
Mean error and root-mean-square error (RMSE) describe their direction and
magnitude, while intervals are graded by reference containment, distinct
from coverage of an expected value. Policy choice is assessed by score
regret, the best reference value minus that of the selected candidate.
For the incumbent, the logged mean supplies its selection estimate, and
independent references supply its grading value. All runs are
retained in these analyses, including extremes.

The configuration and planned analyses were fixed before execution and
are supplied with the code. The cash analyses were specified after an
exploratory comparison had ranked the field visit first under cash and
before the simulations reported here. In contrast, spread,
replicate-consistency checks, top-two decompositions, selection tests and
differences-in-differences are post hoc. For the selection tests, paired
seed-block mean regret differences are compared against zero, with Holm
adjustment across the twelve $t$-tests of each
objective~\citep{holm1979simple}. Of these, nine compare pilot-IS, pilot-DR
and boosted DR with tabular DR, pooling over all conditions, controls
only or coupled conditions only. The remaining three compare tabular DR's
regret with coupling on against off, separately in A, in B and over both
worlds.

\subsection{Estimators and evaluation measures}

The panel contains six log-based and two exploration-based estimators
(Table~\ref{tab:panel}). Family statements exclude context-blind DM.

\begin{table}[!htbp]
  \centering
  \caption{Estimators evaluated in JUBILEE.}
  \label{tab:panel}
  \footnotesize
  \begin{tabular}{lll}
    \toprule
    estimator & construction & reference \\
    \midrule
    DM & tabular reward model, self-evaluated & \citep{dudik2011doubly} \\
    DM (context-blind) & reward model on actions only & this paper \\
    IS & inverse-propensity weighting & \citep{horvitz1952generalization} \\
    SNIPS & self-normalized IS & \citep{swaminathan2015self} \\
    DR & doubly robust, 2-fold cross-fitted model & \citep{dudik2011doubly, chernozhukov2018double} \\
    DM\textsuperscript{+}-IS & DR with annotated reward model & \citep{mandyam2024candor} \\
    \midrule
    pilot-IS & Horvitz--Thompson on explore rows & \citep{horvitz1952generalization} \\
    pilot-DR & doubly robust on explore rows & \citep{dudik2011doubly} \\
    \bottomrule
  \end{tabular}
\end{table}

The direct-method estimator follows \citet{dudik2011doubly}, written as:

\begin{equation}
  \hat V_{\mathrm{DM}} = \tfrac{1}{n}\textstyle\sum_{i}\sum_{a}
    \pi_e(a \mid x_i)\, \hat R(x_i, a) \label{eq:dm}
\end{equation}

where $\hat V_{\mathrm{DM}}$ is the policy-value estimate over $n$ cohort
rows indexed by $i$. The observed context is $x_i$, $a$ ranges over available
actions, $\pi_e(a\mid x_i)$ is the target action probability and
$\hat R(x_i,a)$ the predicted score. The importance-sampling
estimator~\citep{horvitz1952generalization,dudik2011doubly} is given by:

\begin{equation}
  \hat V_{\mathrm{IS}} = \tfrac{1}{n}\textstyle\sum_{i} \rho_i\, r_i
    \label{eq:is}
\end{equation}

where $r_i$ is the logged score for action $a_i$, and
$\rho_i=\pi_e(a_i\mid x_i)/\hat P(a_i\mid x_i)$ is its importance
weight. The denominator $\hat P(a_i\mid x_i)$ estimates the logging
probability from observed context. The cohort size $n$ and contexts
$x_i$ follow Equation~\eqref{eq:dm}. SNIPS normalizes by the sum of
weights~\citep{swaminathan2015self}.
Combining model predictions with weighted residuals gives the doubly
robust estimator~\citep{dudik2011doubly}, written as:

\begin{equation}
  \hat V_{\mathrm{DR}} = \tfrac{1}{n}\textstyle\sum_{i}\Big[
    \sum_{a} \pi_e(a \mid x_i)\, \hat R(x_i, a)
    + \rho_i\big(r_i - \hat R(x_i, a_i)\big)\Big] \label{eq:dr}
\end{equation}

where $r_i-\hat R(x_i,a_i)$ is the prediction residual for the logged
action. The other symbols retain their definitions from
Equations~\eqref{eq:dm} and~\eqref{eq:is}. The estimator combines the DM
prediction average with the importance-weighted residual correction.
Its guarantees require identification and support, which hidden
confounding can invalidate. In the tabular construction, empirical
propensities and a shared 36-by-45 reward grid make DM, IS and SNIPS
algebraically identical when every occupied context supports the target.
With these propensities the importance-weighted model term equals the DM
term for any reward model, so the DR correction vanishes and DR reduces to
IS whatever the reward model, including the annotation-augmented model of
Equation~\eqref{eq:annot}. Cross-fitting and empty-cell handling explain
the small differences between methods.

Pilot estimators use the decisions randomized by the exploration
component, whose assignment probabilities are known. Following the design
in Equation~\eqref{eq:floor}, both use weight $44/\varepsilon$ only on
flagged randomized rows matching the target and retain the full cohort
denominator. Pilot-IS weights observed scores, whereas pilot-DR weights
prediction residuals and also averages model predictions over the full
cohort.

The annotation-augmented reward-model fit follows
CANDOR~\citep{mandyam2024candor}, using logged and weighted annotation
rows as:

\begin{equation}
  \hat R^{+} \;=\; \text{fit on}\;\;
  \underbrace{\{(x_i,\, a_i,\, r_i)\}}_{\text{logged rows}}
  \;\cup\;
  \underbrace{\{(x,\, a^{\pi_e}(x),\, \tilde r_x)\}}_{\text{annotations, weight } \omega}
  \label{eq:annot}
\end{equation}

where $\hat R^{+}$ is the annotation-augmented reward model. The union
combines logged context--action--score triplets with annotation rows.
For an annotation, $x$ is its context, $a^{\pi_e}(x)$ the target action,
$\tilde r_x$ its annotated score and $\omega$ its fitting weight.
Both studies draw annotations from an independent population and add
bias and Gaussian noise at 0.25 of the maximum absolute cell reward, with
unit annotation weight. The auxiliary bandit uses a perfect annotator. Annotations
enter only the reward-model fit in Equation~\eqref{eq:dr}, so under the
tabular propensities above they cannot move the DM\textsuperscript{+}-IS
estimate beyond cross-fitting noise and unsupported cells. CANDOR's bias
guarantee for this construction requires exact importance ratios, which
only the auxiliary bandit has. The annotated direct method
DM\textsuperscript{+} averages $\hat R^{+}$ over the cohort as in
Equation~\eqref{eq:dm}, without the residual correction, so annotations
act on it directly.

\subsection{Gradient-boosted nuisances and inference}

The paired study evaluates DM, IS, SNIPS, DR and pilot-DR with histogram
gradient boosting implemented in scikit-learn~\citep{pedregosa2011scikit}.
Reward models use the action and 15 observed loan and borrower attributes. Target-specific classifiers estimate matching-action
probabilities. Both nuisances use two-fold cross-fitting
\citep{chernozhukov2018double}. Tabular models use coarse contexts and
full-cohort empirical propensities, cross-fitting only the reward model.
The comparison therefore changes covariates and fitting alongside model class.

A reference-free development rule compares the library default with a
regularized configuration by the largest deviation of the mean IS weight
from unity across targets, keeps the smaller deviation and prefers the
default in a near tie. The regularized configuration was selected, with
300 iterations, learning rate 0.05, at most 15 leaves, at least 1,000 rows
per leaf and unit $L_2$ regularization. Propensities are clipped at 0.001,
below the design floor $\varepsilon/44$, and
Appendix~\ref{sec:campaignextra} reports support, clipped counts and mean
weights. Pilot-IS is shared across regimes, and annotation augmentation is
tabular only.

Bootstrap resampling follows \citet{efron1979bootstrap}. Both designs use
nominal 90\% percentile intervals from 200 borrower resamples, conditional
on regional shocks. Fixed-model DM resamples predictions, omitting fitting
uncertainty. IS/SNIPS resample weighted outcomes, DR/augmented DR resample
fixed-nuisance contributions, and pilots use design-weighted contributions.
The DM diagnostic refits its model in each resample. Reference uncertainty
is separate. Auxiliary exploration intervals instead use Student intervals for
seed-level means.

All the experiments presented in this section have their computational
costs summarized here. The grid used 140.8 CPU-hours, about 44 minutes
per run. Reference sizing and the paired study used 17.4 and 94.2 CPU-hours,
the latter including estimation under both objectives. Untimed supporting
diagnostics are estimated at 10--20 CPU-hours, separately from measured
auxiliary costs in Appendix~\ref{sec:scoreequivalence}.

\section{Results}\label{sec:results}

All runs specified in Section~\ref{sec:protocol} have results reported here.
The analysis first checks reference precision and negative controls
(Section~\ref{sec:validationresults}), then compares estimator errors and agreement
under hidden coupling (Section~\ref{sec:frontier}). Estimator precision under randomized exploration
(Section~\ref{sec:price}), model misspecification and interval construction
(Section~\ref{sec:selfeval}), and annotation quality
(Section~\ref{sec:fidelity}) are examined before policy selection and
score regret (Section~\ref{sec:selection}). The final analyses measure
exploration's effect on the incumbent score
(Section~\ref{sec:explorationscore}) and assess rankings and decisions
under simulated cash (Section~\ref{sec:scorecash}).

Auxiliary bandit and annotator experiments are identified separately.
Errors are estimates minus simulated references, reported as means with
one seed-level SE in US\$ per decision. They include reference Monte Carlo
variation. Values and regrets use the score in Equation~\eqref{eq:reward}
unless labeled as cash.

\subsection{Reference precision and negative controls}\label{sec:validationresults}\label{sec:negcontrol}

In the paired study, reference SEs meet the targets in every run, with
maxima of 1.88~US\$ for values and 0.61~US\$ for gaps.
Table~\ref{tab:campaignvalues} averages 20 seeds per condition, sharing
blocks across conditions. The stage policy is reference-best in every
run, with a minimum runner-up gap of 12.2~US\$. The winner is unchanged
when grading uses one, two or three reference populations.

\begin{table}[!htbp]
\centering
\caption{Policy reference scores by condition.}
\label{tab:campaignvalues}
\small
\begin{tabular}{lrrrr}
\toprule
condition & $\pi_b$ & $\pi_{\mathrm{visit}}$ & $\pi_{\mathrm{sms}}$ & $\pi_{\mathrm{stage}}$ \\
\multicolumn{1}{c}{} & \multicolumn{4}{c}{Score (US\$/decision)} \\
\midrule
$A_{\mathrm{off}}$ & 278.9 & 263.5 & 292.2 & 305.4 \\
$A_{\mathrm{on}}$ & 288.0 & 271.0 & 301.1 & 314.3 \\
$B_{\mathrm{off}}$ & 273.6 & 254.4 & 289.3 & 304.6 \\
$B_{\mathrm{on}}$ & 286.4 & 265.3 & 302.4 & 318.0 \\
\bottomrule
\end{tabular}
\end{table}

The incumbent's logged-minus-reference mean ranges from $-0.08$ to
$+0.58$~US\$, with SEs of 0.71--0.83~US\$. The coupling-off conditions serve as negative controls, retaining hidden
outcome variation while disabling its direct influence on channel
assignment. These controls show
no detectable mean error for the context-conditioned tabular family,
boosted DR/SNIPS or pilots. Boosted DM and IS have control errors
(Table~\ref{tab:campaignfullerrors}). Pilot mean-error 95\% intervals
extend 23 to 39~US\$ on either side of zero, leaving smaller biases
unresolved.
A post hoc check finds less reference-mean dispersion than predicted for
values, with unresolved cause (Appendix~\ref{sec:referencechecks}).

The grid's coupling-off control likewise shows no detectable mean error
(Table~\ref{tab:control}, $\varepsilon=0.08$, 20 seeds). Its approximate
80\%-power resolution at the 5\% level is 11 to 16~US\$ for log-based
methods and 37--53~US\$ for pilots, and grid reference SEs are at most
1.97~US\$ per run. Passing this check does not establish zero bias.

\begin{table}[!htbp]
\centering
\caption{Mean estimator errors in the negative control.}
\label{tab:control}
\footnotesize
\begin{tabular}{lrr}
\toprule
estimator & $\pi_{\mathrm{visit}}$ & $\pi_{\mathrm{sms}}$ \\
\multicolumn{1}{c}{} & \multicolumn{2}{c}{Mean error (US\$/decision)} \\
\midrule
log-based five (range of means) & $-3.3$ to $-3.1$ ($\pm 4.0$) & $-0.8$ to $-0.2$ ($\pm 5.8$) \\
DM (context-blind) & $-3.5 \pm 3.7$ & $-1.4 \pm 3.0$ \\
pilot-IS & $+9.3 \pm 13.3$ & $-8.2 \pm 18.5$ \\
pilot-DR & $+5.9 \pm 13.7$ & $-6.4 \pm 19.0$ \\
\bottomrule
\end{tabular}
\end{table}

\subsection{Errors under confounded assignment}\label{sec:frontier}

\begin{figure}[!htbp]
  \centering
  \includegraphics[width=0.74\textwidth]{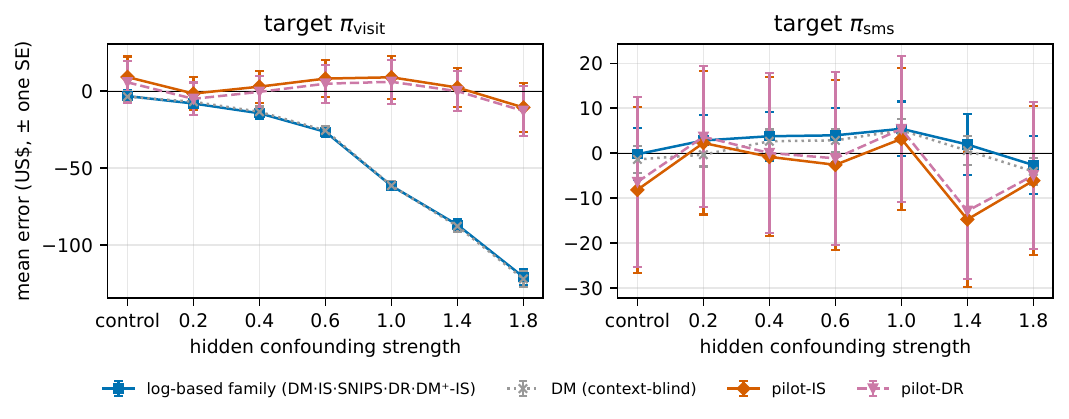}
  \caption{Estimator error across confounding strengths.}
  \label{fig:frontier}
\end{figure}

Across individual grid runs, target-policy reference scores span
203--419~US\$ per decision, with cell means in Table~\ref{tab:gridvalues}.
With coupling enabled, the visit-policy error grows with $\alpha$
(Table~\ref{tab:frontier}, Figure~\ref{fig:frontier}). Each plotted cell
has $\varepsilon=0.08$ and 20 seeds, with bars of one SE. The control
retains $\alpha=0.6$ with coupling off. Five context-conditioned methods
agree within 0.3~US\$ for visits and 0.7~US\$ for SMS, so DR represents
them in the table and one curve represents them in the figure.
Context-blind DM is separate. This agreement follows the tabular algebra
of Section~\ref{sec:protocol} and does not establish accuracy.

Visit errors are $-7.9\pm3.9$ and $-14.2\pm3.8$~US\$ at strengths 0.2
and 0.4, roughly double between 0.4 and 0.6, and reach
$-120.7\pm5.2$~US\$ at 1.8. That is 44\% of the cell's 274.0~US\$
reference value. SMS remains within one SE of zero throughout, using a
tighter plot scale. The noise sweep ($\sigma=0.15$--$0.60$, strength 0.6)
shows no ordered DR-error trend, with interval widths of 60--63~US\$. Its
5-seed cells have SEs of 6 to 8~US\$, so differences between cells below
roughly 30~US\$ are unresolved, limiting generalization about noise.

\begin{table}[!htbp]
\centering
\caption{Mean errors across confounding strengths.}
\label{tab:frontier}
\footnotesize
\begin{adjustbox}{max width=\linewidth}
\begin{tabular}{lrrrrrr}
\toprule
& \multicolumn{3}{c}{$\pi_{\mathrm{visit}}$} & \multicolumn{3}{c}{$\pi_{\mathrm{sms}}$} \\
\cmidrule(lr){2-4}\cmidrule(lr){5-7}
$\alpha$ & DR & pilot-IS & pilot-DR & DR & pilot-IS & pilot-DR \\
\multicolumn{1}{c}{} & \multicolumn{6}{c}{Mean error (US\$/decision)} \\
\midrule
control & $-3.1 \pm 4.0$ & $+9.3 \pm 13.3$ & $+5.9 \pm 13.7$ & $-0.2 \pm 5.8$ & $-8.2 \pm 18.5$ & $-6.4 \pm 19.0$ \\
0.2 & $-7.9 \pm 3.9$ & $-1.4 \pm 10.8$ & $-4.9 \pm 10.6$ & $+2.8 \pm 5.8$ & $+2.3 \pm 15.9$ & $+3.7 \pm 15.7$ \\
0.4 & $-14.2 \pm 3.8$ & $+3.0 \pm 10.4$ & $-0.3 \pm 10.3$ & $+3.8 \pm 5.4$ & $-0.8 \pm 17.7$ & $+0.0 \pm 17.7$ \\
0.6 & $-26.4 \pm 3.1$ & $+8.3 \pm 12.0$ & $+5.0 \pm 12.5$ & $+4.0 \pm 6.1$ & $-2.6 \pm 18.9$ & $-1.1 \pm 19.2$ \\
1.0 & $-61.4 \pm 2.6$ & $+9.1 \pm 14.0$ & $+6.2 \pm 14.3$ & $+5.4 \pm 6.1$ & $+3.2 \pm 15.8$ & $+5.4 \pm 16.3$ \\
1.4 & $-86.8 \pm 3.7$ & $+2.5 \pm 12.8$ & $+0.1 \pm 13.0$ & $+1.9 \pm 6.8$ & $-14.8 \pm 15.0$ & $-12.9 \pm 15.2$ \\
1.8 & $-120.7 \pm 5.2$ & $-10.5 \pm 16.1$ & $-12.7 \pm 16.3$ & $-2.6 \pm 6.5$ & $-6.1 \pm 16.6$ & $-4.9 \pm 16.4$ \\
\bottomrule
\end{tabular}
\end{adjustbox}
\end{table}

The paired study retains a visit-policy underestimation of approximately
one fifth under coupling (Table~\ref{tab:campaignvisit}, mean error
$\pm$ one SE over 20 seeds). Boosted DM, DR and SNIPS share errors of a
similar scale. Boosted IS is less negative because it carries a positive
offset of 9 to 14~US\$ on every target in the controls, where its mean
importance weight exceeds one, so its on--off contrast matches the others.
Boosted DM errs by $-9.3\pm1.5$ and $-14.2\pm1.3$~US\$ on the stage policy
in the controls, an error that the DR correction removes. Pilot errors are
compatible with zero at wider uncertainty, and the RMSE comparison in
Appendix~\ref{sec:campaignextra} shows no uniform pilot advantage.

\begin{table}[!htbp]
\centering
\caption{Field-visit policy score errors.}
\label{tab:campaignvisit}
\small
\begin{adjustbox}{max width=\linewidth}
\begin{tabular}{lrrrr}
\toprule
estimator & $A_{\mathrm{off}}$ & $A_{\mathrm{on}}$ & $B_{\mathrm{off}}$ & $B_{\mathrm{on}}$ \\
\multicolumn{1}{c}{} & \multicolumn{4}{c}{Mean error (US\$/decision)} \\
\midrule
tabular DR & $-1.1 \pm 4.8$ & $-57.5 \pm 3.8$ & $-4.0 \pm 4.7$ & $-56.7 \pm 3.6$ \\
boosted DM & $-4.3 \pm 1.9$ & $-53.0 \pm 1.9$ & $-3.7 \pm 1.9$ & $-50.3 \pm 2.0$ \\
boosted IS & $+14.1 \pm 5.2$ & $-43.9 \pm 3.9$ & $+10.7 \pm 5.0$ & $-42.9 \pm 3.9$ \\
boosted SNIPS & $-0.7 \pm 4.9$ & $-53.5 \pm 3.7$ & $-3.5 \pm 4.7$ & $-52.3 \pm 3.8$ \\
boosted DR & $-3.9 \pm 5.1$ & $-55.3 \pm 3.9$ & $-7.3 \pm 4.8$ & $-53.8 \pm 4.0$ \\
pilot-IS & $-5.3 \pm 13.8$ & $-4.9 \pm 14.1$ & $-6.9 \pm 13.6$ & $-8.6 \pm 11.5$ \\
tabular pilot-DR & $-4.6 \pm 13.1$ & $-4.9 \pm 13.6$ & $-6.2 \pm 12.8$ & $-8.6 \pm 11.0$ \\
\bottomrule
\end{tabular}
\end{adjustbox}
\end{table}

\begin{figure}[!htbp]
\centering
\includegraphics[width=\linewidth]{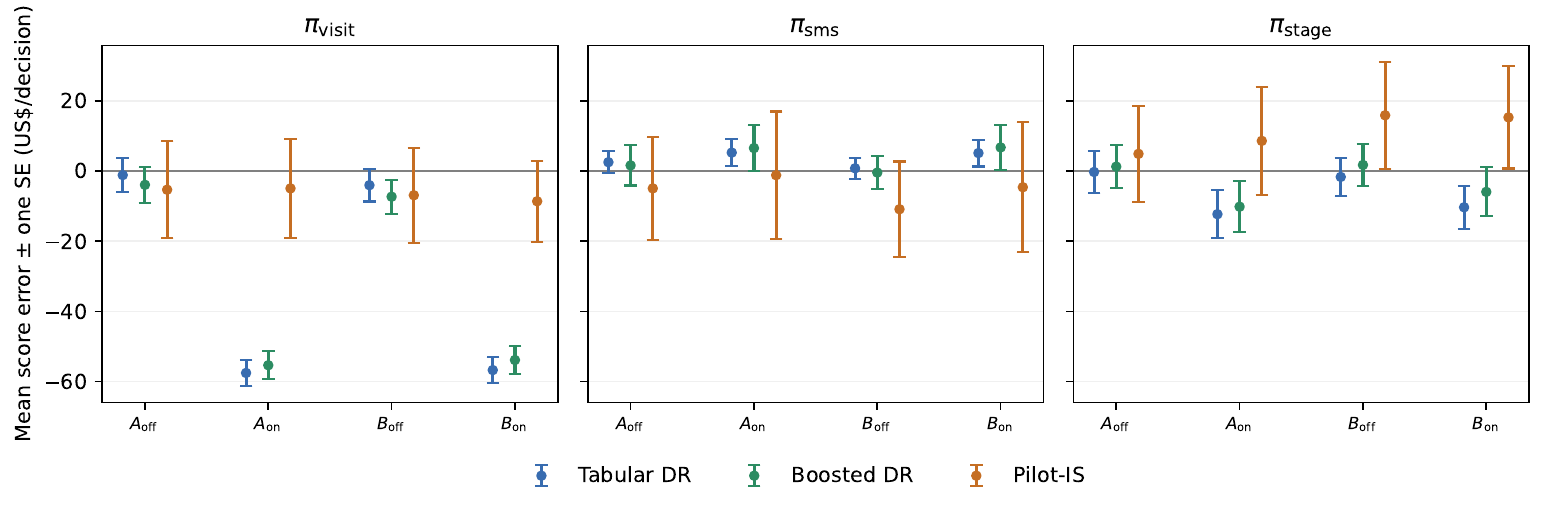}
\caption{Estimator errors by policy and assignment condition.}
\label{fig:campaignerrors}
\end{figure}

An exploratory within-run spread, the largest minus smallest log-based
estimate, averages 0.6--2.0~US\$ for tabular methods and 18--29~US\$ for
boosted methods, or 12--25~US\$ without boosted IS, including controls
(Table~\ref{tab:campaignagreement}).
Thus disagreement also occurs without hidden coupling, while coupled
boosted methods can still share downward error. Figure~\ref{fig:campaignerrors}
compares tabular DR, boosted DR and pilot-IS on a common scale, with one-SE
bars over shared seed blocks. Full results and additional paired
contrasts appear in the appendices.

World B moves values and gaps without changing their ordering. Its coupled
tabular-DR visit error differs from A by only $+0.8\pm1.8$~US\$.
Grid cell 06 shares the configuration of condition $A_{\mathrm{on}}$ with
different seeds, and the two agree on the reference scores within
0.4~US\$ and on the tabular DR visit error within one SE, $-61.4\pm2.6$
against $-57.5\pm3.8$~US\$.

\subsection{Identification from randomized exploration}\label{sec:price}

Pilots use flagged randomization rather than estimated logging propensities.
In the control and at every coupled strength of the 20-seed ladder,
neither pilot deviates from zero by more than one SE, while log-based error reaches
$-120.7$~US\$ (Table~\ref{tab:frontier}).

Without exploration the pilots are undefined in five cells. With
$\varepsilon=0.08$, approximately 59{,}000 decisions are randomized, but
only 1{,}277 to 1{,}453 per run match a given target in the grid, and
1{,}279 to 1{,}425 in the paired study. Their visit-policy 90\% intervals are
2.7--3.1 times wider than DR's on the noise ladder, at 168--186~US\$
against 60--63~US\$.

\subsection{Self-evaluation and interval construction}\label{sec:selfeval}

A contextual bandit isolates reward-model misspecification using exact
propensities and a perfect annotator, with target value 0.3327 per unit
of debt, $N=2000$ logged decisions per seed, $\varepsilon=0.1$ and 200 seeds. The well-specified
model conditions on context and action, while the misspecified model uses
actions alone. In Table~\ref{tab:bandit}, all well-specified biases are
within about one Monte Carlo SE of zero. Misspecified DM loses
$0.1476\pm0.0029$, or 44\% of truth, while importance-corrected estimates
remain compatible with zero.

DM\textsuperscript{+}-IS has the lowest RMSE in both regimes. Shared-seed
squared-error comparisons yield paired $t$-statistics of at least 2.7
when well-specified and 3.8 when misspecified ($n=200$ seeds). This is pairwise evidence under
a perfect annotator, not a recommendation for unknown annotation quality.

\begin{table}[!htbp]
  \centering
  \caption{Estimator performance in the contextual bandit.}
  \label{tab:bandit}
  \footnotesize
  \begin{tabular}{lrrrr}
    \toprule
    & \multicolumn{2}{c}{well-specified} & \multicolumn{2}{c}{misspecified} \\
    \cmidrule(lr){2-3}\cmidrule(lr){4-5}
    estimator & bias & RMSE & bias & RMSE \\
    \midrule
    DM (self-eval.) & $+.000 \pm .007$ & $0.097$ & $-.148 \pm .003$ & $0.153$ \\
    IS & $+.008 \pm .009$ & $0.129$ & $+.008 \pm .009$ & $0.129$ \\
    DR (cross-fit.) & $-.001 \pm .008$ & $0.106$ & $+.004 \pm .008$ & $0.110$ \\
    DM\textsuperscript{+}-IS & $+.003 \pm .006$ & $\mathbf{0.085}$ & $+.003 \pm .007$ & $\mathbf{0.095}$ \\
    \bottomrule
  \end{tabular}
\end{table}

The grid also distinguishes point estimates from interval construction
(Table~\ref{tab:coverage}). All methods use the same 165 runs, excluding
25 without exploration. The fractions pool cells of every confounding
strength, so the visit column mixes near-zero and large errors, and
Tables~\ref{tab:controlintervals} and \ref{tab:campaignintervals}
separate control from coupled conditions. Fractions describe nominal
90\% reference containment and widths are in US\$ per decision.
IS/SNIPS contain about
half the visit references and 86\% of SMS references despite point values
close to DM. Fixed-model DM intervals omit fitting uncertainty and fail
even in controls. Context-blind DM intervals are degenerate.

\begin{table}[!htbp]
\centering
\caption{Reference containment and interval width.}
\label{tab:coverage}
\footnotesize
\begin{tabular}{lrrrr}
\toprule
& \multicolumn{2}{c}{$\pi_{\mathrm{visit}}$} & \multicolumn{2}{c}{$\pi_{\mathrm{sms}}$} \\
\cmidrule(lr){2-3}\cmidrule(lr){4-5}
estimator & fraction & mean width & fraction & mean width \\
 & & (US\$/decision) & & (US\$/decision) \\
\midrule
DM (fixed model) & 0.012 & 0.8 & 0.012 & 1.2 \\
DM (context-blind) & 0.000 & 0.0 & 0.000 & 0.0 \\
IS & 0.479 & 59.7 & 0.855 & 69.8 \\
SNIPS & 0.479 & 59.3 & 0.855 & 69.6 \\
DR (cross-fitted) & 0.479 & 59.2 & 0.855 & 70.5 \\
DM\textsuperscript{+}-IS & 0.479 & 60.0 & 0.855 & 70.6 \\
pilot-IS & 0.861 & 179.9 & 0.842 & 208.6 \\
pilot-DR & 0.830 & 178.1 & 0.830 & 206.3 \\
\bottomrule
\end{tabular}
\end{table}

In the 20-run control of Table~\ref{tab:controlintervals}
($\varepsilon=0.08$), DR and refitted DM intervals contain the reference
in 18 or 19 runs, whereas fixed-model DM intervals contain none. Pilots
contain more pooled visit references at about three times the width. Shared seeds prevent treating pooled fractions as
independent binomial observations, and none incorporates reference error.

\begin{table}[!htbp]
\centering
\caption{Intervals in the negative control.}
\label{tab:controlintervals}
\footnotesize
\begin{tabular}{lrrrr}
\toprule
 & \multicolumn{2}{c}{$\pi_{\mathrm{visit}}$} & \multicolumn{2}{c}{$\pi_{\mathrm{sms}}$} \\
\cmidrule(lr){2-3}\cmidrule(lr){4-5}
estimator & fraction & mean width & fraction & mean width \\
 &  & (US\$/decision) &  & (US\$/decision) \\
\midrule
DM (fixed model) & 0.000 & 1.0 & 0.000 & 1.2 \\
DM refit & 0.950 & 73.3 & 0.900 & 69.5 \\
DR & 0.950 & 73.4 & 0.900 & 71.3 \\
pilot-IS & 0.800 & 187.2 & 0.850 & 209.9 \\
\bottomrule
\end{tabular}
\end{table}

Refitting tabular DM within bootstrap resamples widens intervals
(Table~\ref{tab:campaignintervals}, containment counts out of 20 with
mean full widths in parentheses). Control widths imply bootstrap SDs
close to across-seed error SDs for visits and stage policies, and
30--35\% larger for SMS, with only 19 degrees of freedom. Refit intervals
contain control references in 18--19 runs, but coupled visit references
in only 2 per world. Fixed-nuisance DR intervals have similar widths and
the same containment pattern, 17 to 19 of 20 in the controls and 2 of 20
for the coupled visit policy. Twenty runs per
condition cannot establish precise nominal coverage.

\begin{table}[!htbp]
\centering
\caption{Interval containment and width in the paired study.}
\label{tab:campaignintervals}
\small
\begin{adjustbox}{max width=\linewidth}
\begin{tabular}{llrrrr}
\toprule
condition & target & DM (fixed model) & DM refit & DR & pilot-IS \\
\multicolumn{2}{c}{} & \multicolumn{4}{c}{Contained / runs (mean width in US\$/decision)} \\
\midrule
$A_{\mathrm{off}}$ & $\pi_{\mathrm{visit}}$ & 0/20 (1.0) & 19/20 (70.8) & 18/20 (71.5) & 18/20 (184.6) \\
$A_{\mathrm{off}}$ & $\pi_{\mathrm{sms}}$ & 2/20 (1.1) & 19/20 (61.7) & 19/20 (62.5) & 18/20 (214.5) \\
$A_{\mathrm{off}}$ & $\pi_{\mathrm{stage}}$ & 0/20 (1.2) & 18/20 (86.1) & 17/20 (86.2) & 18/20 (223.4) \\
$B_{\mathrm{off}}$ & $\pi_{\mathrm{visit}}$ & 1/20 (0.9) & 18/20 (69.6) & 17/20 (70.1) & 18/20 (181.3) \\
$B_{\mathrm{off}}$ & $\pi_{\mathrm{sms}}$ & 2/20 (1.1) & 19/20 (59.2) & 19/20 (60.4) & 18/20 (208.8) \\
$B_{\mathrm{off}}$ & $\pi_{\mathrm{stage}}$ & 0/20 (1.1) & 18/20 (85.7) & 18/20 (85.7) & 18/20 (225.6) \\
$A_{\mathrm{on}}$ & $\pi_{\mathrm{visit}}$ & 0/20 (0.8) & 2/20 (58.6) & 2/20 (58.7) & 17/20 (186.3) \\
$B_{\mathrm{on}}$ & $\pi_{\mathrm{visit}}$ & 0/20 (0.8) & 2/20 (58.8) & 2/20 (59.2) & 19/20 (184.1) \\
\bottomrule
\end{tabular}
\end{adjustbox}
\end{table}

\subsection{Annotator fidelity}\label{sec:fidelity}

Under tabular empirical propensities, the importance-weighted correction
cancels the reward model (Section~\ref{sec:protocol}), and
DM\textsuperscript{+}-IS stays within 0.3~US\$ of IS in the paired study
whatever the annotations.

To examine annotation quality without this cancellation, the annotated
direct method DM\textsuperscript{+} is evaluated with exact and corrupted
annotations (Table~\ref{tab:dmplus}, field-visit policy, score).
Exact annotations are context means of the target policy's own outcomes
in the independent annotation population, one per logged row.
With exact annotations, DM\textsuperscript{+}'s error lies between $-0.8$
and $+0.1$~US\$ for every condition, target and objective, including the coupled
visit policy whose log-based error is $-57.5\pm3.8$~US\$.

With the specified corruption, whose bias is 0.25 of the largest cell
reward, DM\textsuperscript{+}'s error instead rises to $+150$ to
$+243$~US\$.

\begin{table}[!htbp]
\centering
\caption{Annotated estimators for the field-visit policy.}
\label{tab:dmplus}
\footnotesize
\begin{adjustbox}{max width=\linewidth}
\begin{tabular}{lrrrr}
\toprule
condition & DM, logs only & DM\textsuperscript{+}, exact & DM\textsuperscript{+}, corrupted & DM\textsuperscript{+}-IS, corrupted \\
\multicolumn{1}{c}{} & \multicolumn{4}{c}{Mean error (US\$/decision)} \\
\midrule
$A_{\mathrm{off}}$ & $-1.1 \pm 4.8$ & $-0.2 \pm 0.7$ & $+208.2 \pm 22.7$ & $-1.0 \pm 4.8$ \\
$A_{\mathrm{on}}$ & $-57.5 \pm 3.8$ & $-0.8 \pm 0.7$ & $+212.6 \pm 22.4$ & $-57.4 \pm 3.8$ \\
$B_{\mathrm{off}}$ & $-3.9 \pm 4.6$ & $-0.0 \pm 0.6$ & $+203.4 \pm 22.9$ & $-3.9 \pm 4.6$ \\
$B_{\mathrm{on}}$ & $-56.6 \pm 3.6$ & $-0.6 \pm 0.6$ & $+212.6 \pm 21.9$ & $-56.5 \pm 3.6$ \\
\bottomrule
\end{tabular}
\end{adjustbox}
\end{table}

\subsection{Policy selection and score regret}\label{sec:selection}

A post hoc grid analysis selects the highest estimated value among the
incumbent and two targets, grading against realized references on the
same 165 runs. Table~\ref{tab:selection} reports correct counts and mean
score regret in US\$ per decision. SMS is reference-best in all 190 runs.
Because the same candidate wins every run, a constant choice would have
zero regret, and the counts measure how often estimation error displaces
a fixed winner. Context-conditioned log-based methods have lower observed
regret than pilots, and context-blind DM has the lowest.
Configuration-level values and choices appear in Tables~\ref{tab:gridvalues} and
\ref{tab:cellselection}.

\begin{table}[!htbp]
\centering
\caption{Policy selection in the parameter grid.}
\label{tab:selection}
\footnotesize
\begin{tabular}{lrrr}
\toprule
estimator & correct / runs & percent & mean regret \\
 & & & (US\$/decision) \\
\midrule
DM & 108 / 165 & 65.5 & 4.64 \\
DM (context-blind) & 145 / 165 & 87.9 & 1.50 \\
IS & 108 / 165 & 65.5 & 4.64 \\
SNIPS & 108 / 165 & 65.5 & 4.64 \\
DR (cross-fitted) & 106 / 165 & 64.2 & 4.79 \\
DM\textsuperscript{+}-IS & 108 / 165 & 65.5 & 4.64 \\
pilot-IS & 86 / 165 & 52.1 & 10.32 \\
pilot-DR & 87 / 165 & 52.7 & 9.73 \\
\bottomrule
\end{tabular}
\end{table}

The paired study adds the stage challenger, which is reference-best in
every run. All five context-conditioned tabular methods make identical
choices. Table~\ref{tab:campaignselection} gives correct selections out
of 20 and mean score regret $\pm$ one SE in parentheses, comparing their
DR representative with boosted DR and tabular pilots. Pilots have higher
observed regret in every condition. Most incorrect log-based choices
select SMS. Selection errors also occur in controls.

\begin{table}[!htbp]
\centering
\caption{Policy selection in the paired study.}
\label{tab:campaignselection}
\small
\begin{adjustbox}{max width=\linewidth}
\begin{tabular}{lrrrr}
\toprule
condition & tabular DR & boosted DR & pilot-IS & pilot-DR \\
\multicolumn{1}{c}{} & \multicolumn{4}{c}{Correct / runs (mean regret in US\$/decision)} \\
\midrule
$A_{\mathrm{off}}$ & 12/20 ($6.7 \pm 2.3$) & 13/20 ($7.3 \pm 2.7$) & 6/20 ($16.1 \pm 3.3$) & 7/20 ($14.6 \pm 3.3$) \\
$A_{\mathrm{on}}$ & 8/20 ($8.6 \pm 1.7$) & 8/20 ($9.9 \pm 2.1$) & 6/20 ($14.9 \pm 3.1$) & 8/20 ($14.2 \pm 3.3$) \\
$B_{\mathrm{off}}$ & 15/20 ($3.9 \pm 1.6$) & 16/20 ($4.9 \pm 2.7$) & 12/20 ($10.2 \pm 3.5$) & 13/20 ($7.8 \pm 2.9$) \\
$B_{\mathrm{on}}$ & 10/20 ($7.8 \pm 1.8$) & 11/20 ($7.1 \pm 1.8$) & 10/20 ($10.5 \pm 3.0$) & 11/20 ($8.8 \pm 2.8$) \\
\bottomrule
\end{tabular}
\end{adjustbox}
\end{table}

An exploratory comparison helps explain these selection errors. Under
coupling, the stage-minus-SMS error is $-17.5\pm7.3$~US\$ in A and
$-15.5\pm7.4$~US\$ in B, against reference gaps of 13.2 and 15.6~US\$.
The expected estimated gap is therefore close to zero, and correct
selections fall from 12 and 15 of 20 in the controls to 8 and 10.
Twenty blocks leave this displacement unresolved as a reversal.

The regret tests likewise limit the selection conclusions. Under the
score, none of the twelve post hoc tests survives Holm adjustment, with
a minimum adjusted $p=0.25$, so a general regret advantage is
unestablished. Reference precision has little effect on the reported
mean regret. Using fewer reference populations changes mean regret by at
most 0.14~US\$, while retaining the winner. True rankings remain fixed
within each study.

\subsection{Exploration and the score objective}\label{sec:explorationscore}

A fixed-hazard comparison changes incumbent exploration from
$\varepsilon=0$ to $0.08$ throughout the nine-month window, unlike the
intervention defined in Section~\ref{sec:intervention}. Each of 20 shared seed blocks uses a paired
million-borrower population, with sizing and precision rules fixed in
development. Table~\ref{tab:explorationscore} gives
$\Delta=V(\varepsilon=0)-V(\varepsilon=0.08)$ and 95\% $t$ intervals
across seeds, plus mean relative reductions. Here $V(\varepsilon)$ is
the expected incumbent score under exploration share $\varepsilon$.
Positive differences in both worlds indicate a lower score under exploration.
The score-equivalence result in Appendix~\ref{sec:scoreequivalence} applies
to this comparison. Full-settlement cash effects of exploration have not
been evaluated.

\begin{table}[!htbp]
\centering
\caption{Effect of randomized exploration on the score.}
\label{tab:explorationscore}
\small
\begin{tabular}{lrr}
\toprule
world & $\Delta$ [95\% interval] & relative reduction \\
 & (US\$/decision) & \\
\midrule
$A_{\mathrm{on}}$ & $+1.66\;[+1.41, +1.92]$ & 0.57\% \\
$B_{\mathrm{on}}$ & $+1.41\;[+1.12, +1.70]$ & 0.49\% \\
\bottomrule
\end{tabular}
\end{table}

\subsection{Score and simulated cash}\label{sec:scorecash}

Because policy rankings depend on the objective, the cash analysis
examines whether estimation errors lead to choosing candidates that
recover less money in the simulator. Both studies repeat every estimate,
interval and selection on the same simulations using the cash objective
defined in Section~\ref{sec:scoredefinition}.

The field-visit policy becomes reference-best in all 80 paired runs
(Table~\ref{tab:campaignvalues_cash}) and in all 190 grid runs. In the
paired study, its mean margin is 5.8 to 15.9~US\$ per decision, while
reference SEs reach at most 1.31~US\$ for values and 0.44~US\$ for
paired gaps. Cash also excludes channel costs.

\begin{table}[!htbp]
\centering
\caption{Policy reference cash values by condition.}
\label{tab:campaignvalues_cash}
\footnotesize
\begin{tabular}{lrrrr}
\toprule
condition & $\pi_b$ & $\pi_{\mathrm{visit}}$ & $\pi_{\mathrm{sms}}$ & $\pi_{\mathrm{stage}}$ \\
\multicolumn{1}{c}{} & \multicolumn{4}{c}{Cash value (US\$/decision)} \\
\midrule
$A_{\mathrm{off}}$ & 206.0 & 222.0 & 205.6 & 206.2 \\
$A_{\mathrm{on}}$ & 208.5 & 225.8 & 209.3 & 209.9 \\
$B_{\mathrm{off}}$ & 203.1 & 214.4 & 205.4 & 208.6 \\
$B_{\mathrm{on}}$ & 207.9 & 220.6 & 211.6 & 214.8 \\
\bottomrule
\end{tabular}
\end{table}

Despite the new ranking, the visit-policy error persists. Under coupling,
tabular DR's cash errors are $-49.6\pm3.5$ and $-49.1\pm3.2$~US\$ in A
and B, about 22\% of its value. Boosted DR's errors are $-47.3\pm3.8$ and
$-46.4\pm3.7$~US\$, while pilot errors remain compatible with zero
(Appendix~\ref{sec:cashextra}). Tabular DR intervals again contain the
coupled visit reference in 2 of 20 runs per world.

The corresponding policy choices appear in
Table~\ref{tab:campaignselection_cash}. Coupling
raises tabular DR's cash regret by $6.3\pm1.2$~US\$ per decision over both
worlds, with Holm-adjusted $p<0.001$, and lowers its correct selections
from 11 and 8 of 20 to 1 and 0.

\begin{table}[!htbp]
\centering
\caption{Policy selection in the paired study under cash.}
\label{tab:campaignselection_cash}
\footnotesize
\begin{adjustbox}{max width=\linewidth}
\begin{tabular}{llrrrr}
\toprule
condition & reference best & tabular DR & boosted DR & pilot-IS & pilot-DR \\
\multicolumn{2}{c}{} & \multicolumn{4}{c}{Correct / runs (mean regret in US\$/decision)} \\
\midrule
$A_{\mathrm{off}}$ & $\pi_{\mathrm{visit}}$ & 11/20 ($7.3 \pm 1.8$) & 10/20 ($8.3 \pm 1.9$) & 7/20 ($10.7 \pm 1.8$) & 7/20 ($10.7 \pm 1.8$) \\
$A_{\mathrm{on}}$ & $\pi_{\mathrm{visit}}$ & 1/20 ($15.8 \pm 0.9$) & 1/20 ($15.8 \pm 0.9$) & 6/20 ($11.7 \pm 1.8$) & 7/20 ($10.9 \pm 1.9$) \\
$B_{\mathrm{off}}$ & $\pi_{\mathrm{visit}}$ & 8/20 ($4.5 \pm 0.9$) & 7/20 ($5.2 \pm 1.0$) & 2/20 ($7.7 \pm 0.8$) & 4/20 ($6.6 \pm 0.9$) \\
$B_{\mathrm{on}}$ & $\pi_{\mathrm{visit}}$ & 0/20 ($8.5 \pm 0.5$) & 1/20 ($8.3 \pm 0.7$) & 4/20 ($7.1 \pm 1.0$) & 6/20 ($6.2 \pm 1.0$) \\
\bottomrule
\end{tabular}
\end{adjustbox}
\end{table}

Pilots select the visit policy more often, in 4 to 7 of 20 coupled runs.
Their observed regret is 2.7 and 3.6~US\$ below tabular DR's for pilot-IS
and pilot-DR, but these differences do not survive Holm adjustment
(adjusted $p=0.33$ and 0.19). The grid shows a similar selection pattern.
Log-based estimators select the cash-best candidate in 33 or 34 of the
165 runs with exploration, none of them at strengths of 1.0 and above,
and pilots in 66 or 68. Appendix~\ref{sec:appendix} reports the
confounding ladder and selection under cash.

\section{Discussion and Limitations}\label{sec:discussion}

JUBILEE quantifies OPE failure modes under the specified collections
mechanisms. Tabular agreement coexists with large errors, while boosted
methods disagree in both controls and coupled conditions. Neither pattern
alone establishes accuracy.

The larger visit-policy error is consistent with the assignment mechanism
in Section~\ref{sec:floor}. Higher hidden friction and lower
contactability reduce recovery while increasing assignment intensity,
making visits more likely for borrowers with adverse characteristics
unavailable to log-based estimators. This selection helps explain the
downward visit error. SMS has no direct intensity effect on its channel
weight, although normalization can still induce confounding. The smaller
SMS error is specific to the tested settings, and the different candidate
discounts preclude attributing the entire contrast to channel selection.

Counterfactual annotations offer additional information for reward
prediction. DM\textsuperscript{+}-IS trains its model on logged outcomes
and annotations, then corrects its value estimate with importance-weighted
differences between observed and predicted rewards. With correct ratios,
identification and support, this correction offsets model error in
expectation, as in conventional DR~\citep{dudik2011doubly,mandyam2024candor}.
Better predictions can reduce variability, although imperfect annotations
can increase it. The auxiliary bandit's exact propensities and perfect
annotator provide favorable conditions for the lower RMSE observed in both
model specifications (Section~\ref{sec:selfeval}).

In the coupled collections experiments, however, observed-covariate
propensities lack that exact-ratio guarantee. The tabular empirical ratios
also cancel the model terms, leaving DM\textsuperscript{+}-IS nearly
identical to IS even with annotations (Section~\ref{sec:fidelity}).
DM\textsuperscript{+}, which uses predictions without residual correction,
instead inherits the accuracy or bias of the annotations dominating its
fit. Its near-reference result with exact annotations reflects privileged
target-policy outcomes, and corrupted annotations transfer bias to its
predictions. Thus annotation quality and estimator construction jointly
determine the benefit. Because these annotations use controlled simulator
outcomes, human annotator performance remains untested. The contrast
between the bandit and collections shows that robustness to reward-model
misspecification does not imply robustness to hidden confounding.

Known randomization supplies an alternative source of information, as in
microrandomized trials~\citep{klasnja2015microrandomized}. Pilots avoid
the large visit error at the experiment's resolution, but rare rewards
and few matching randomized decisions limit precision. This limit and the
approximation from binning continuous offers leave smaller pilot biases
unresolved.

These precision limits carry into policy selection
(Section~\ref{sec:selection}). Selection depends on relative estimates,
so shared shifts can preserve rankings while differential errors can
change a close comparison. Under the score, the largest error penalizes
an already inferior visit policy, whereas the stage--SMS error erodes
the leading candidate's advantage. Pilot variability can also change the
winner, explaining why smaller mean error need not improve selection.
Fixed reference winners further restrict the range of decision changes
tested by these designs.

Randomization also changes the incumbent's score
(Section~\ref{sec:explorationscore}). The observed reduction combines
changes in channels and offers under the score's initial-discount
convention, without isolating their individual contributions. Since the
comparison excludes information value and has not been repeated for
full-settlement cash, it does not establish the economic cost of
exploration.

The cash comparison makes the objective's role in policy selection
explicit (Section~\ref{sec:scorecash}). For loans unrecovered initially
under both candidates, common continuation produces identical subsequent
cash trajectories. The score still weights these recoveries by each
candidate's initial discount, favoring lower offers. Under simulated cash,
visits become the best candidate, so the shared downward error now makes
log-based estimators discard the winner. The increase in tabular DR's cash
regret under coupling is resolved, whereas regret differences between
pilots and log-based estimators remain unresolved under both objectives. These contrasts
support reporting error, uncertainty and choices together without a
universal estimator recommendation.

Interpreting these comparisons also requires distinguishing uncertainty
in model fitting from uncertainty in grading. The DM refit diagnostic
exposes omitted fitting variability while retaining coupled point error,
without invalidating all fixed-nuisance DR intervals. Separately,
independent reference populations separate grading from logs and
annotations, and pairing improves gap precision, but their SEs remain
conditional on regional shocks.

Generalization is further limited by the tested mechanisms. The worlds
vary one channel-effect mechanism with shared seed blocks, while latent
weights, distributions, alternative clipping and bootstrap budgets lack
broad sensitivity analysis. The coupling switch also changes observed
assignment, support and continuation, while grid hazard recalibration
changes the outcome model alongside each stress parameter. These contrasts
therefore do not isolate universal confounding thresholds. Likewise, the
boosted comparison changes covariates and propensity cross-fitting
alongside model class, preventing separate attribution. The panel also
excludes confounding-aware and sensitivity methods~\citep{namkoong2020offpolicy,
brunssmith2021causality,kausik2024offline,hess2026efficient}.

Finally, the absence of external portfolios and the omission of channel
costs and borrower welfare limit operational interpretation. Contact
intensity toward distressed borrowers is also subject to consumer-protection
rules, adding constraints beyond the recovery outcomes evaluated here.
The supported scope is therefore a synthetic initial-action evaluation,
without operational collections recommendations or full sequential-policy
claims. Within this scope, JUBILEE connects estimator errors to information
conditions and policy choices through reproducible experiments.

\section{Conclusion}\label{sec:conclusion}

JUBILEE provides a reproducible environment for investigating when OPE
failure modes arise, quantifying their magnitude and assessing their
effects on policy choice under specified collections mechanisms.

The experiments show that estimator agreement can coexist with substantial
error. Assessing reliability therefore requires examining uncertainty as
well as point estimates, including the fitting variability omitted by
fixed-model DM intervals. This uncertainty also affects policy choice.
Estimates based on randomized decisions do not show the large
underestimation observed in the coupled logs, but their variability across
runs limits their ability to distinguish candidates. A smaller error on
average therefore does not necessarily lead to more accurate policy
selection in individual runs.

The evaluation objective further determines how estimation error affects
that choice. Under the synthetic score, the visit policy is not the best
candidate. Under simulated cash, it becomes the best, yet log-based
estimators continue to underestimate its value. This underestimation can
make another candidate appear preferable and lead to the selection of a
policy with lower simulated cash recovery.

The simulator and evaluation protocol are available
\ifdefined\JubileeAnonymous in the accompanying implementation\else at
\url{https://github.com/alpha-synthesis/jubilee}, and the cohort extracts of
every run at \url{https://doi.org/10.5281/zenodo.24008544}\fi.

\ifdefined\JubileeAnonymous
\section*{LLM statement}
\else
\section*{Acknowledgments}

The authors review all AI-assisted contributions and take responsibility
for the manuscript, code and reported results.

\subsection*{LLM statement}
\fi

Large language models assisted with software implementation, analysis
checks and manuscript revision. LLM review agents helped check claims
and numerical consistency against saved simulation outputs.
\ifdefined\JubileeAnonymous
The authors review all AI-assisted contributions and take responsibility
for the manuscript, code and reported results.
\fi

Loan records, rewards and policy-value estimates are produced by the
simulator and evaluation code using the specified configurations and
random seeds. These outputs are not edited by LLMs under any
circumstances. Reported statistics are computed by code from saved
inputs and outputs.

\bibliographystyle{tmlr}
\bibliography{paper}

@inproceedings{dudik2011doubly,
  author    = {Miroslav Dud\'{i}k and John Langford and Lihong Li},
  title     = {Doubly Robust Policy Evaluation and Learning},
  booktitle = {Proceedings of ICML},
  year      = {2011},
  pages     = {1097--1104},
  url       = {https://arxiv.org/abs/1103.4601}
}

@article{horvitz1952generalization,
  author  = {Daniel G. Horvitz and Donovan J. Thompson},
  title   = {A Generalization of Sampling Without Replacement from a Finite Universe},
  journal = {Journal of the American Statistical Association},
  volume  = {47},
  number  = {260},
  pages   = {663--685},
  year    = {1952},
  doi     = {10.1080/01621459.1952.10483446}
}

@inproceedings{swaminathan2015self,
  author    = {Adith Swaminathan and Thorsten Joachims},
  title     = {The Self-Normalized Estimator for Counterfactual Learning},
  booktitle = {Advances in Neural Information Processing Systems},
  volume    = {28},
  year      = {2015},
  url       = {https://papers.nips.cc/paper_files/paper/2015/hash/39027dfad5138c9ca0c474d71db915c3-Abstract.html}
}

@article{chernozhukov2018double,
  author  = {Victor Chernozhukov and Denis Chetverikov and Mert Demirer and Esther Duflo and Christian Hansen and Whitney Newey and James Robins},
  title   = {Double/debiased machine learning for treatment and structural parameters},
  journal = {The Econometrics Journal},
  volume  = {21},
  number  = {1},
  pages   = {C1--C68},
  year    = {2018},
  doi     = {10.1111/ectj.12097}
}

@inproceedings{mandyam2024candor,
  author    = {Aishwarya Mandyam and Shengpu Tang and Jiayu Yao and Jenna Wiens and Barbara E. Engelhardt},
  title     = {{CANDOR}: Counterfactual ANnotated DOubly Robust Off-Policy Evaluation},
  booktitle = {Proceedings of CHIL},
  series    = {Proceedings of Machine Learning Research},
  volume    = {333},
  pages     = {644--687},
  publisher = {PMLR},
  year      = {2026},
  url       = {https://proceedings.mlr.press/v333/mandyam26a.html}
}

@inproceedings{tang2023isplus,
  author    = {Shengpu Tang and Jenna Wiens},
  title     = {Counterfactual-Augmented Importance Sampling for Semi-Offline Policy Evaluation},
  booktitle = {Advances in Neural Information Processing Systems},
  volume    = {36},
  year      = {2023},
  url       = {https://arxiv.org/abs/2310.17146}
}

@inproceedings{oberst2019counterfactual,
  author    = {Michael Oberst and David Sontag},
  title     = {Counterfactual Off-Policy Evaluation with {Gumbel-Max} Structural Causal Models},
  booktitle = {Proceedings of ICML},
  series    = {Proceedings of Machine Learning Research},
  volume    = {97},
  pages     = {4881--4890},
  year      = {2019},
  publisher = {PMLR},
  url       = {https://proceedings.mlr.press/v97/oberst19a.html}
}

@inproceedings{namkoong2020offpolicy,
  author    = {Hongseok Namkoong and Ramtin Keramati and Steve Yadlowsky and Emma Brunskill},
  title     = {Off-policy Policy Evaluation For Sequential Decisions Under Unobserved Confounding},
  booktitle = {Advances in Neural Information Processing Systems},
  volume    = {33},
  year      = {2020},
  url       = {https://arxiv.org/abs/2003.05623}
}

@inproceedings{voloshin2021cobs,
  author    = {Voloshin, Cameron and Le, Hoang M. and Jiang, Nan and Yue, Yisong},
  title     = {Empirical Study of Off-Policy Policy Evaluation for Reinforcement Learning},
  booktitle = {NeurIPS Datasets and Benchmarks},
  year      = {2021},
  url       = {https://arxiv.org/abs/1911.06854}
}

@inproceedings{fu2021dope,
  author    = {Fu, Justin and Norouzi, Mohammad and Nachum, Ofir and Tucker, George and Wang, Ziyu and Novikov, Alexander and Yang, Mengjiao and Zhang, Michael R. and Chen, Yutian and Kumar, Aviral and Paduraru, Cosmin and Levine, Sergey and Paine, Tom Le},
  title     = {Benchmarks for Deep Off-Policy Evaluation},
  booktitle = {Proceedings of ICLR},
  year      = {2021},
  url       = {https://arxiv.org/abs/2103.16596}
}

@inproceedings{saito2021obp,
  author    = {Saito, Yuta and Aihara, Shunsuke and Matsutani, Megumi and Narita, Yusuke},
  title     = {{Open Bandit Dataset} and Pipeline: Towards Realistic and Reproducible Off-Policy Evaluation},
  booktitle = {NeurIPS Datasets and Benchmarks},
  year      = {2021},
  url       = {https://arxiv.org/abs/2008.07146}
}

@misc{kiyohara2023scoperl,
  author    = {Kiyohara, Haruka and Kishimoto, Ren and Kawakami, Kosuke and Kobayashi, Ken and Nakata, Kazuhide and Saito, Yuta},
  title     = {{SCOPE-RL}: A {Python} Library for Offline Reinforcement Learning and Off-Policy Evaluation},
  year      = {2023},
  doi       = {10.48550/arXiv.2311.18206},
  howpublished = {arXiv preprint arXiv:2311.18206},
  url       = {https://arxiv.org/abs/2311.18206}
}

@inproceedings{byrd2020abides,
  author    = {Byrd, David and Hybinette, Maria and Balch, Tucker Hybinette},
  title     = {{ABIDES}: Towards High-Fidelity Multi-Agent Market Simulation},
  booktitle = {Proc. ACM SIGSIM Conf. Principles of Advanced Discrete Simulation (SIGSIM-PADS '20)},
  pages     = {11--22},
  year      = {2020},
  doi       = {10.1145/3384441.3395986}
}

@inproceedings{assefa2020generating,
  author    = {Assefa, Samuel A. and Dervovic, Danial and Mahfouz, Mahmoud and Tillman, Robert E. and Reddy, Prashant and Veloso, Manuela},
  title     = {Generating Synthetic Data in Finance: Opportunities, Challenges and Pitfalls},
  booktitle = {Proc. First ACM Int. Conf. AI in Finance (ICAIF '20)},
  year      = {2020},
  doi       = {10.1145/3383455.3422554}
}

@article{kriebel2020,
  author    = {Kriebel, Johannes and Yam, Kevin},
  title     = {Forecasting Recoveries in Debt Collection: Debt Collectors and Information Production},
  journal   = {European Financial Management},
  volume    = {26},
  number    = {3},
  pages     = {537--559},
  year      = {2020},
  doi       = {10.1111/eufm.12242}
}

@inproceedings{collectionsrl2025,
  author    = {Brand{\~a}o, Bruno and Martins, Luana Guedes Barros and de Oliveira, Bryan Lincoln Marques and Melo, Luckeciano Carvalho and da Luz, Murilo Lopes and Garcia, Eduardo and da Silva, Marcos Vinicius and Avelar, Renato Gnecco and Galv{\~a}o Filho, Arlindo Rodrigues and Soares, Anderson da Silva and Soares, Telma Woerle de Lima},
  title     = {Reinforcement Learning for Debt Pricing: A Case Study in Financial Services},
  booktitle = {RLC Workshop on RL4RS},
  year      = {2025},
  url       = {https://openreview.net/forum?id=1cfG46owm8}
}

@inproceedings{li2011unbiased,
  author = {Li, Lihong and Chu, Wei and Langford, John and Wang, Xuanhui},
  title = {Unbiased Offline Evaluation of Contextual-Bandit-Based News Article Recommendation Algorithms},
  booktitle = {Proc. Fourth ACM Int. Conf. Web Search and Data Mining (WSDM '11)},
  year = {2011},
  pages = {297--306},
  doi = {10.1145/1935826.1935878}
}

@article{klasnja2015microrandomized,
  author = {Klasnja, Predrag and Hekler, Eric B. and Shiffman, Saul and Boruvka, Audrey and Almirall, Daniel and Tewari, Ambuj and Murphy, Susan A.},
  title = {Microrandomized Trials: An Experimental Design for Developing Just-in-Time Adaptive Interventions},
  journal = {Health Psychology},
  year = {2015},
  volume = {34},
  number = {Suppl},
  pages = {1220--1228},
  doi = {10.1037/hea0000305}
}

@misc{cfpb2021creditcard,
  author      = {{Consumer Financial Protection Bureau}},
  title       = {The Consumer Credit Card Market},
  year        = {2021},
  url         = {https://files.consumerfinance.gov/f/documents/cfpb_consumer-credit-card-market-report_2021.pdf}
}

@misc{ftc2013debtbuying,
  author      = {{Federal Trade Commission}},
  title       = {The Structure and Practices of the Debt Buying Industry},
  year        = {2013},
  url         = {https://www.ftc.gov/reports/structure-practices-debt-buying-industry}
}

@inproceedings{brunssmith2021causality,
  author    = {Bruns-Smith, David A.},
  title     = {Model-Free and Model-Based Policy Evaluation when Causality is Uncertain},
  booktitle = {Proceedings of ICML},
  series    = {Proceedings of Machine Learning Research},
  volume    = {139},
  pages     = {1116--1126},
  year      = {2021},
  publisher = {PMLR},
  url       = {https://proceedings.mlr.press/v139/bruns-smith21a.html}
}

@inproceedings{kausik2024offline,
  author    = {Kausik, Chinmaya and Lu, Yangyi and Tan, Kevin and Makar, Maggie and Wang, Yixin and Tewari, Ambuj},
  title     = {Offline Policy Evaluation and Optimization Under Confounding},
  booktitle = {Proceedings of the 27th International Conference on Artificial Intelligence and Statistics},
  series    = {Proceedings of Machine Learning Research},
  volume    = {238},
  pages     = {1459--1467},
  year      = {2024},
  publisher = {PMLR},
  url       = {https://proceedings.mlr.press/v238/kausik24a.html}
}

@inproceedings{hess2026efficient,
  author    = {Hess, Konstantin and Frauen, Dennis and Melnychuk, Valentyn and Feuerriegel, Stefan},
  title     = {Efficient and Sharp Off-Policy Learning under Unobserved Confounding},
  year      = {2026},
  booktitle = {Proceedings of ICLR},
  doi       = {10.48550/arXiv.2502.13022},
  url       = {https://proceedings.iclr.cc/paper_files/paper/2026/file/5e972fe492946fc7a02aef64d0f4fd7e-Paper-Conference.pdf}
}

@inproceedings{sanzmartin2026adaptive,
  author    = {Sanz-Mart{\'i}n, Laura and Monroy-Tepepa, Joysiane and Jammal, Manal and Verdasco-S{\'a}nchez, Victoria and Zato-Dom{\'i}nguez, Carolina},
  title     = {Adaptive Intelligence in Debt Collection: Enhancing Financial Resilience via Reinforcement Learning},
  booktitle = {Proceedings of the 2026 5th International Conference on Cyber Security, Artificial Intelligence and Digital Economy (CSAIDE)},
  pages     = {286--291},
  year      = {2026},
  publisher = {Association for Computing Machinery},
  doi       = {10.1145/3819836.3819879}
}

@misc{ycenycollections,
  author    = {Tracy Yang},
  title     = {{RL-project-debt-collection}},
  howpublished = {GitHub repository},
  year      = {2019},
  url       = {https://github.com/yceny/RL-project-debt-collection},
  note      = {Accessed November 5, 2026}
}

@article{holm1979simple,
  author  = {Sture Holm},
  title   = {A Simple Sequentially Rejective Multiple Test Procedure},
  journal = {Scandinavian Journal of Statistics},
  volume  = {6},
  number  = {2},
  pages   = {65--70},
  year    = {1979},
  url     = {https://www.jstor.org/stable/4615733}
}

@article{efron1979bootstrap,
  author  = {Bradley Efron},
  title   = {Bootstrap Methods: Another Look at the Jackknife},
  journal = {The Annals of Statistics},
  volume  = {7},
  number  = {1},
  pages   = {1--26},
  year    = {1979},
  doi     = {10.1214/aos/1176344552}
}

@article{pedregosa2011scikit,
  author  = {Fabian Pedregosa and Ga{{\"e}}l Varoquaux and Alexandre Gramfort and Vincent Michel and Bertrand Thirion and Olivier Grisel and Mathieu Blondel and Peter Prettenhofer and Ron Weiss and Vincent Dubourg and Jake Vanderplas and Alexandre Passos and David Cournapeau and Matthieu Brucher and Matthieu Perrot and {{\'E}}douard Duchesnay},
  title   = {{Scikit-learn}: Machine Learning in {Python}},
  journal = {Journal of Machine Learning Research},
  year    = {2011},
  volume  = {12},
  number  = {85},
  pages   = {2825--2830},
  url     = {https://jmlr.org/papers/v12/pedregosa11a.html}
}

@incollection{mcfadden1974conditional,
  author = {Daniel McFadden},
  title = {Conditional Logit Analysis of Qualitative Choice Behavior},
  booktitle = {Frontiers in Econometrics},
  editor = {P. Zarembka},
  pages = {105--142},
  publisher = {Academic Press},
  address = {New York},
  year = {1974},
  url = {https://eml.berkeley.edu/reprints/mcfadden/zarembka.pdf}
}

@misc{nist2003handbook,
  author = {{NIST/SEMATECH}},
  title = {e-Handbook of Statistical Methods},
  year = {2003},
  howpublished = {Sections 1.3.6.2 and 1.3.6.6.17},
  url = {https://www.itl.nist.gov/div898/handbook/eda/section3/eda366h.htm},
  note = {Accessed November 7, 2026}
}
\clearpage
\appendix
\section{Configuration and reference values}\label{sec:appendix}

The synthetic model combines the world-A channel effects and base behavior
weights in Table~\ref{tab:coefficients} with an observable-risk term,
written as:

\begin{equation}
\begin{aligned}
f(z_i) ={}& -0.0009((\mathrm{age}_i-43)/10)^2 + 0.07 \\
 &-0.018\,\mathrm{dpd}_i/30 - 0.05\,L_i/30 \\
 &-0.012\,T_i/12
\end{aligned}
\end{equation}

where $f(z_i)$ is the observable contribution to recovery log-odds for
loan $i$. Its covariates $z_i$ comprise borrower age $\mathrm{age}_i$ in
years, days past due $\mathrm{dpd}_i$, late-stage duration $L_i$ in days
and loan age $T_i$ in months. The coefficients and scaling constants are
synthetic settings without portfolio fitting. The discount response in
Equation~\eqref{eq:hazard} uses $\beta=1.1$ and $\kappa=3$.

\begin{table}[!htbp]
\centering
\caption{Channel effects and assignment parameters.}
\label{tab:coefficients}
\footnotesize
\begin{tabular}{lrrr}
\toprule
channel & log-odds effect & base weight $w_c$ & intensity coefficient $\eta_c$ \\
\midrule
none & $-0.85$ & 0.20 & $-0.55$ \\
SMS & $+0.10$ & 0.25 & $0$ \\
app message & $+0.18$ & 0.25 & $0$ \\
phone & $+0.35$ & 0.20 & $+0.35$ \\
field visit & $+0.48$ & 0.10 & $+0.55$ \\
\bottomrule
\end{tabular}
\end{table}

For assignment, Table~\ref{tab:coefficients} gives the intensity
coefficients with coupling enabled. Setting them to zero recovers the
coupling-off channel weights in Equation~\eqref{eq:channelpolicy}.
The incumbent's offer before rounding is
$D_i=\min\{d_{\max},d_{\max}U_i+\delta_i\}$ with
$U_i\sim\mathrm{Beta}(2,5)$, where $d_{\max}=0.55$ is the offer cap and
$\delta_i=\max(0,\mathrm{dpd}_i-60)/360$ is the delinquency adjustment.
With bin width $h=0.05$, the first ten intervals are $I_k=[kh,(k+1)h)$,
and $I_{10}=[10h,d_{\max}]$ includes the cap's probability mass.
Using this offer law and the standard Beta cumulative distribution
function~\citep{nist2003handbook}, the nominal probabilities of
Equation~\eqref{eq:discountbins} are given by:

\begin{equation}
q(k\mid\mathrm{dpd}_i)=
\begin{cases}
F\!\left(\dfrac{(k+1)h-\delta_i}{d_{\max}}\right)
-F\!\left(\dfrac{kh-\delta_i}{d_{\max}}\right), & k=0,\ldots,9,\\[7pt]
1-F\!\left(\dfrac{10h-\delta_i}{d_{\max}}\right), & k=10
\end{cases}
\label{eq:discountcdf}
\end{equation}

where $F$ is the cumulative distribution function of a Beta$(2,5)$ draw,
extended by zero below its support and one above it.

The implementation rounds offers to four decimal places. The formulas
in Section~\ref{sec:floor} describe nominal probabilities before this
rounding, including the uniform exploration mass $\varepsilon/44$.
Reported estimates and references use the generated offers.

The 17 grid configurations are listed in Table~\ref{tab:gridvalues}, together
with calibrated monthly hazards $\tau$, run counts, realized annual
recovery, mean reference scores in US\$ per decision and the largest
reference SE of each cell.
The control has $\alpha=0.6$ with coupling off. Incumbent values include
exploration.

\begin{table}[!htbp]
\centering
\caption{Parameter configurations, realized recovery and policy reference scores.}
\label{tab:gridvalues}
\footnotesize
\begin{adjustbox}{max width=\linewidth}
\begin{tabular}{crcrrrrrrrrr}
\toprule
cell & $\alpha$ & coupling & $\sigma$ & $\varepsilon$ & $\tau$ & runs & recovery (\%) & behavior & $\pi_{\mathrm{visit}}$ & $\pi_{\mathrm{sms}}$ & max ref. SE \\
\multicolumn{8}{c}{} & \multicolumn{4}{c}{Reference values and SE (US\$/decision)} \\
\midrule
01 & 0.6 & off & 0.25 & 0.00 & 0.001882 & 5 & 4.28 & 293.7 & 275.7 & 305.3 & 1.73 \\
02 & 0.6 & off & 0.25 & 0.08 & 0.001885 & 20 & 4.27 & 290.6 & 274.9 & 304.3 & 1.77 \\
03 & 0.6 & on & 0.25 & 0.00 & 0.001700 & 5 & 4.26 & 288.3 & 269.8 & 300.8 & 1.71 \\
04 & 0.6 & on & 0.25 & 0.08 & 0.001700 & 20 & 4.23 & 283.8 & 267.7 & 298.3 & 1.75 \\
05 & 1.0 & on & 0.25 & 0.00 & 0.001461 & 5 & 4.38 & 300.0 & 280.3 & 311.7 & 1.78 \\
06 & 1.0 & on & 0.25 & 0.08 & 0.001421 & 20 & 4.23 & 287.6 & 270.7 & 300.9 & 1.83 \\
07 & 1.4 & on & 0.25 & 0.00 & 0.001015 & 5 & 3.98 & 276.9 & 258.7 & 286.4 & 1.72 \\
08 & 1.4 & on & 0.25 & 0.08 & 0.000992 & 20 & 3.88 & 267.5 & 251.7 & 278.7 & 1.83 \\
09 & 1.8 & on & 0.25 & 0.00 & 0.000752 & 5 & 4.13 & 293.5 & 273.4 & 302.4 & 1.79 \\
10 & 1.8 & on & 0.25 & 0.08 & 0.000756 & 20 & 4.15 & 292.3 & 274.0 & 303.3 & 1.97 \\
11 & 0.6 & on & 0.15 & 0.08 & 0.001700 & 5 & 4.19 & 279.9 & 264.0 & 294.3 & 1.70 \\
12 & 0.6 & on & 0.40 & 0.08 & 0.001558 & 5 & 4.10 & 275.0 & 259.2 & 289.1 & 1.68 \\
13 & 0.6 & on & 0.60 & 0.08 & 0.001443 & 5 & 4.19 & 281.0 & 264.8 & 295.4 & 1.70 \\
14 & 1.4 & on & 0.40 & 0.08 & 0.000950 & 5 & 3.94 & 271.9 & 255.7 & 283.5 & 1.72 \\
15 & 1.8 & on & 0.60 & 0.08 & 0.000648 & 5 & 4.16 & 292.5 & 273.9 & 303.6 & 1.80 \\
16 & 0.2 & on & 0.25 & 0.08 & 0.001700 & 20 & 3.87 & 258.1 & 243.7 & 272.0 & 1.61 \\
17 & 0.4 & on & 0.25 & 0.08 & 0.001700 & 20 & 4.00 & 267.2 & 252.1 & 281.2 & 1.67 \\
\bottomrule
\end{tabular}
\end{adjustbox}
\end{table}

For these configurations, Table~\ref{tab:cellselection} reports correct choices
and mean score regret
in parentheses, using each cell's run count as denominator. ``--'' means
undefined. Cells share seeds, and the common comparison excludes
configurations without exploration. The bandit and annotation diagnostics
are separate from the 190-run aggregates.

\begin{table}[!htbp]
\centering
\caption{Policy selection by parameter configuration.}
\label{tab:cellselection}
\footnotesize
\begin{adjustbox}{max width=\linewidth}
\begin{tabular}{crrrrrrrr}
\toprule
cell & DM & DM-blind & IS & SNIPS & DR & DM\textsuperscript{+}-IS & pilot-IS & pilot-DR \\
\multicolumn{1}{c}{} & \multicolumn{8}{c}{Correct choices (mean regret in US\$/decision)} \\
\midrule
01 & 4 (5.54) & 4 (5.54) & 2 (10.31) & 4 (5.54) & 4 (5.54) & 5 (0.00) & -- & -- \\
02 & 11 (7.00) & 16 (2.71) & 11 (7.00) & 11 (7.00) & 11 (7.00) & 11 (7.00) & 8 (13.14) & 9 (11.68) \\
03 & 4 (2.56) & 5 (0.00) & 3 (5.12) & 4 (2.56) & 4 (2.56) & 5 (0.00) & -- & -- \\
04 & 15 (3.61) & 19 (0.70) & 15 (3.61) & 15 (3.61) & 15 (3.61) & 15 (3.61) & 10 (11.36) & 10 (11.36) \\
05 & 5 (0.00) & 5 (0.00) & 3 (4.73) & 5 (0.00) & 5 (0.00) & 5 (0.00) & -- & -- \\
06 & 13 (4.77) & 19 (0.62) & 13 (4.77) & 13 (4.77) & 13 (4.77) & 13 (4.77) & 11 (11.19) & 12 (8.90) \\
07 & 5 (0.00) & 5 (0.00) & 4 (1.91) & 5 (0.00) & 5 (0.00) & 5 (0.00) & -- & -- \\
08 & 11 (5.18) & 17 (1.82) & 11 (5.18) & 11 (5.18) & 11 (5.18) & 11 (5.18) & 9 (10.23) & 10 (9.10) \\
09 & 5 (0.00) & 5 (0.00) & 1 (7.35) & 5 (0.00) & 5 (0.00) & 5 (0.00) & -- & -- \\
10 & 13 (4.09) & 14 (3.03) & 13 (4.09) & 13 (4.09) & 12 (4.60) & 13 (4.09) & 10 (10.11) & 9 (11.82) \\
11 & 4 (2.89) & 5 (0.00) & 4 (2.89) & 4 (2.89) & 4 (2.89) & 4 (2.89) & 3 (8.93) & 3 (5.57) \\
12 & 3 (5.47) & 4 (2.82) & 3 (5.47) & 3 (5.47) & 3 (5.47) & 3 (5.47) & 3 (8.83) & 3 (5.47) \\
13 & 3 (5.63) & 4 (2.90) & 3 (5.63) & 3 (5.63) & 3 (5.63) & 3 (5.63) & 3 (9.10) & 3 (9.10) \\
14 & 4 (2.31) & 5 (0.00) & 4 (2.31) & 4 (2.31) & 4 (2.31) & 4 (2.31) & 3 (4.54) & 3 (4.54) \\
15 & 4 (2.23) & 5 (0.00) & 4 (2.23) & 4 (2.23) & 4 (2.23) & 4 (2.23) & 4 (2.23) & 4 (2.23) \\
16 & 14 (4.11) & 18 (1.38) & 14 (4.11) & 14 (4.11) & 13 (4.80) & 14 (4.11) & 11 (10.57) & 11 (9.85) \\
17 & 13 (4.92) & 19 (0.68) & 13 (4.92) & 13 (4.92) & 13 (4.92) & 13 (4.92) & 11 (10.13) & 10 (10.85) \\
\bottomrule
\end{tabular}
\end{adjustbox}
\end{table}

Every grid run is graded by three independent million-borrower reference
populations, as in the paired study. Under simulated cash,
Tables~\ref{tab:frontier_cash} and \ref{tab:selection_cash} repeat the
confounding ladder and the selection analysis. The visit policy is
reference-best in every run, and log-based errors keep the sign and growth
of the score's.

\begin{table}[!htbp]
\centering
\caption{Mean errors across confounding strengths under cash.}
\label{tab:frontier_cash}
\footnotesize
\begin{adjustbox}{max width=\linewidth}
\begin{tabular}{lrrrrrr}
\toprule
& \multicolumn{3}{c}{$\pi_{\mathrm{visit}}$} & \multicolumn{3}{c}{$\pi_{\mathrm{sms}}$} \\
\cmidrule(lr){2-4}\cmidrule(lr){5-7}
$\alpha$ & DR & pilot-IS & pilot-DR & DR & pilot-IS & pilot-DR \\
\multicolumn{1}{c}{} & \multicolumn{6}{c}{Mean error (US\$/decision)} \\
\midrule
control & $-3.7 \pm 3.7$ & $+7.8 \pm 11.7$ & $+4.9 \pm 12.1$ & $+1.7 \pm 4.3$ & $-1.7 \pm 13.7$ & $-0.5 \pm 14.2$ \\
0.2 & $-8.2 \pm 3.4$ & $-3.6 \pm 9.2$ & $-6.4 \pm 9.1$ & $+2.7 \pm 3.6$ & $+2.2 \pm 11.1$ & $+2.9 \pm 10.8$ \\
0.4 & $-12.8 \pm 3.3$ & $+1.9 \pm 9.1$ & $-0.9 \pm 9.1$ & $+3.6 \pm 3.4$ & $+1.8 \pm 12.3$ & $+2.3 \pm 12.3$ \\
0.6 & $-23.4 \pm 2.7$ & $+6.2 \pm 10.4$ & $+3.5 \pm 10.8$ & $+4.2 \pm 3.8$ & $+2.2 \pm 13.0$ & $+3.2 \pm 13.1$ \\
1.0 & $-53.6 \pm 2.5$ & $+8.4 \pm 13.2$ & $+6.1 \pm 13.5$ & $+4.9 \pm 4.1$ & $+3.6 \pm 11.6$ & $+5.3 \pm 12.0$ \\
1.4 & $-75.2 \pm 3.2$ & $+5.7 \pm 12.7$ & $+3.9 \pm 12.9$ & $+2.3 \pm 4.5$ & $-10.1 \pm 10.2$ & $-8.6 \pm 10.4$ \\
1.8 & $-105.7 \pm 4.6$ & $-6.8 \pm 16.1$ & $-8.5 \pm 16.2$ & $-0.6 \pm 4.6$ & $-1.4 \pm 12.8$ & $-0.3 \pm 12.8$ \\
\bottomrule
\end{tabular}
\end{adjustbox}
\end{table}

\begin{table}[!htbp]
\centering
\caption{Policy selection in the parameter grid under cash.}
\label{tab:selection_cash}
\footnotesize
\begin{tabular}{lrrr}
\toprule
estimator & correct / runs & percent & mean regret \\
 & & & (US\$/decision) \\
\midrule
DM & 33 / 165 & 20.0 & 13.16 \\
DM (context-blind) & 34 / 165 & 20.6 & 12.81 \\
IS & 33 / 165 & 20.0 & 13.16 \\
SNIPS & 33 / 165 & 20.0 & 13.16 \\
DR (cross-fitted) & 34 / 165 & 20.6 & 13.14 \\
DM\textsuperscript{+}-IS & 34 / 165 & 20.6 & 13.08 \\
pilot-IS & 68 / 165 & 41.2 & 9.60 \\
pilot-DR & 66 / 165 & 40.0 & 9.78 \\
\bottomrule
\end{tabular}
\end{table}

\section{Complete results for the paired study}\label{sec:campaignextra}
The complete comparisons in Tables~\ref{tab:campaignfullerrors},
\ref{tab:campaignfullerrorsA_on}, \ref{tab:campaignfullerrorsB_off} and
\ref{tab:campaignfullerrorsB_on} cover every fitted estimator in both regimes.
They report mean score errors and seed-level standard errors in US\$
per decision against independent simulated references, using 20 seeds
per condition. Comparing regimes requires accounting for their joint
changes in covariates, model class and propensity cross-fitting.
Pilot-IS is shared across regimes, and shared seed blocks remain the
unit for paired comparisons.

Complementing these error comparisons, Table~\ref{tab:campaignagreement}
reports within-regime disagreement among log-based methods. The exploratory spread is the
largest minus smallest estimate within a run, in US\$ per decision,
averaged over 20 blocks per condition. For the tabular regime it includes
DM, IS, SNIPS, DR and augmented DR, while the boosted comparison includes
DM, IS, SNIPS and DR. The third row of each condition repeats the boosted
spread without IS, whose positive control offset accounts for much of the
disagreement.

\begin{table}[!htbp]
\centering
\caption{Estimator disagreement by model regime.}
\label{tab:campaignagreement}
\small
\begin{tabular}{llrrr}
\toprule
condition & regime & $\pi_{\mathrm{visit}}$ & $\pi_{\mathrm{sms}}$ & $\pi_{\mathrm{stage}}$ \\
\multicolumn{2}{c}{} & \multicolumn{3}{c}{Mean spread (US\$/decision)} \\
\midrule
$A_{\mathrm{off}}$ & tabular & 0.6 & 2.0 & 0.7 \\
$A_{\mathrm{off}}$ & boosted & 26.2 & 24.1 & 25.8 \\
$A_{\mathrm{off}}$ & boosted without IS & 15.5 & 19.0 & 22.2 \\
$A_{\mathrm{on}}$ & tabular & 0.6 & 1.7 & 1.0 \\
$A_{\mathrm{on}}$ & boosted & 19.3 & 25.3 & 25.3 \\
$A_{\mathrm{on}}$ & boosted without IS & 13.4 & 19.4 & 24.9 \\
$B_{\mathrm{off}}$ & tabular & 0.6 & 1.7 & 0.7 \\
$B_{\mathrm{off}}$ & boosted & 25.2 & 21.9 & 29.1 \\
$B_{\mathrm{off}}$ & boosted without IS & 16.5 & 16.9 & 25.3 \\
$B_{\mathrm{on}}$ & tabular & 0.6 & 1.7 & 1.1 \\
$B_{\mathrm{on}}$ & boosted & 17.7 & 25.1 & 25.1 \\
$B_{\mathrm{on}}$ & boosted without IS & 12.4 & 19.6 & 24.5 \\
\bottomrule
\end{tabular}
\end{table}

\begin{table}[!htbp]
\centering
\caption{Estimator score errors in condition $A_{\mathrm{off}}$.}
\label{tab:campaignfullerrors}
\small
\begin{adjustbox}{max width=\linewidth}
\begin{tabular}{llrrr}
\toprule
regime & estimator & $\pi_{\mathrm{visit}}$ & $\pi_{\mathrm{sms}}$ & $\pi_{\mathrm{stage}}$ \\
\multicolumn{2}{c}{} & \multicolumn{3}{c}{Mean error (US\$/decision)} \\
\midrule
tabular & DM & $-1.1 \pm 4.8$ & $+2.0 \pm 3.1$ & $-0.1 \pm 6.1$ \\
tabular & DM, context-blind & $-1.5 \pm 4.1$ & $+7.4 \pm 2.7$ & $+2.6 \pm 5.2$ \\
tabular & IS & $-1.2 \pm 4.8$ & $+1.9 \pm 3.1$ & $-0.3 \pm 6.1$ \\
tabular & SNIPS & $-1.1 \pm 4.8$ & $+2.0 \pm 3.1$ & $-0.1 \pm 6.1$ \\
tabular & DR & $-1.1 \pm 4.8$ & $+2.5 \pm 3.2$ & $-0.3 \pm 6.1$ \\
tabular & DM\textsuperscript{+}-IS & $-1.0 \pm 4.8$ & $+2.2 \pm 3.1$ & $-0.0 \pm 6.1$ \\
tabular & pilot-IS & $-5.3 \pm 13.8$ & $-4.9 \pm 14.6$ & $+4.9 \pm 13.6$ \\
tabular & pilot-DR & $-4.6 \pm 13.1$ & $-3.6 \pm 13.6$ & $+2.7 \pm 12.4$ \\
boosted & DM & $-4.3 \pm 1.9$ & $+1.9 \pm 1.3$ & $-9.3 \pm 1.5$ \\
boosted & IS & $+14.1 \pm 5.2$ & $+11.4 \pm 5.8$ & $+8.6 \pm 6.0$ \\
boosted & SNIPS & $-0.7 \pm 4.9$ & $+1.6 \pm 5.7$ & $+2.4 \pm 5.8$ \\
boosted & DR & $-3.9 \pm 5.1$ & $+1.6 \pm 5.7$ & $+1.3 \pm 6.1$ \\
boosted & pilot-DR & $-5.9 \pm 13.7$ & $-3.3 \pm 14.0$ & $+2.4 \pm 13.1$ \\
\bottomrule
\end{tabular}
\end{adjustbox}
\end{table}

\begin{table}[!htbp]
\centering
\caption{Estimator score errors in condition $A_{\mathrm{on}}$.}
\label{tab:campaignfullerrorsA_on}
\small
\begin{adjustbox}{max width=\linewidth}
\begin{tabular}{llrrr}
\toprule
regime & estimator & $\pi_{\mathrm{visit}}$ & $\pi_{\mathrm{sms}}$ & $\pi_{\mathrm{stage}}$ \\
\multicolumn{2}{c}{} & \multicolumn{3}{c}{Mean error (US\$/decision)} \\
\midrule
tabular & DM & $-57.5 \pm 3.8$ & $+4.9 \pm 3.7$ & $-12.6 \pm 6.7$ \\
tabular & DM, context-blind & $-58.8 \pm 3.2$ & $+8.7 \pm 4.1$ & $-9.4 \pm 5.6$ \\
tabular & IS & $-57.7 \pm 3.8$ & $+4.8 \pm 3.8$ & $-12.8 \pm 6.7$ \\
tabular & SNIPS & $-57.5 \pm 3.8$ & $+4.9 \pm 3.8$ & $-12.6 \pm 6.7$ \\
tabular & DR & $-57.5 \pm 3.8$ & $+5.3 \pm 3.9$ & $-12.3 \pm 6.8$ \\
tabular & DM\textsuperscript{+}-IS & $-57.4 \pm 3.8$ & $+5.1 \pm 3.8$ & $-12.5 \pm 6.7$ \\
tabular & pilot-IS & $-4.9 \pm 14.1$ & $-1.2 \pm 18.1$ & $+8.6 \pm 15.4$ \\
tabular & pilot-DR & $-4.9 \pm 13.6$ & $+0.1 \pm 17.3$ & $+6.0 \pm 14.5$ \\
boosted & DM & $-53.0 \pm 1.9$ & $+3.0 \pm 1.9$ & $-7.2 \pm 2.6$ \\
boosted & IS & $-43.9 \pm 3.9$ & $+17.3 \pm 6.8$ & $-8.4 \pm 6.8$ \\
boosted & SNIPS & $-53.5 \pm 3.7$ & $+6.6 \pm 6.6$ & $-10.3 \pm 6.8$ \\
boosted & DR & $-55.3 \pm 3.9$ & $+6.5 \pm 6.5$ & $-10.2 \pm 7.2$ \\
boosted & pilot-DR & $-4.8 \pm 13.8$ & $+0.7 \pm 17.4$ & $+6.7 \pm 14.9$ \\
\bottomrule
\end{tabular}
\end{adjustbox}
\end{table}

\begin{table}[!htbp]
\centering
\caption{Estimator score errors in condition $B_{\mathrm{off}}$.}
\label{tab:campaignfullerrorsB_off}
\small
\begin{adjustbox}{max width=\linewidth}
\begin{tabular}{llrrr}
\toprule
regime & estimator & $\pi_{\mathrm{visit}}$ & $\pi_{\mathrm{sms}}$ & $\pi_{\mathrm{stage}}$ \\
\multicolumn{2}{c}{} & \multicolumn{3}{c}{Mean error (US\$/decision)} \\
\midrule
tabular & DM & $-3.9 \pm 4.6$ & $+0.6 \pm 3.1$ & $-1.7 \pm 5.5$ \\
tabular & DM, context-blind & $-2.7 \pm 4.1$ & $+9.2 \pm 2.5$ & $-0.4 \pm 4.7$ \\
tabular & IS & $-4.1 \pm 4.6$ & $+0.4 \pm 3.1$ & $-1.8 \pm 5.5$ \\
tabular & SNIPS & $-4.0 \pm 4.6$ & $+0.6 \pm 3.1$ & $-1.6 \pm 5.5$ \\
tabular & DR & $-4.0 \pm 4.7$ & $+0.8 \pm 3.1$ & $-1.7 \pm 5.5$ \\
tabular & DM\textsuperscript{+}-IS & $-3.9 \pm 4.6$ & $+0.7 \pm 3.1$ & $-1.6 \pm 5.5$ \\
tabular & pilot-IS & $-6.9 \pm 13.6$ & $-10.9 \pm 13.6$ & $+15.9 \pm 15.3$ \\
tabular & pilot-DR & $-6.2 \pm 12.8$ & $-10.0 \pm 12.6$ & $+13.9 \pm 13.9$ \\
boosted & DM & $-3.7 \pm 1.9$ & $+1.4 \pm 1.1$ & $-14.2 \pm 1.3$ \\
boosted & IS & $+10.7 \pm 5.0$ & $+9.4 \pm 4.8$ & $+9.0 \pm 6.0$ \\
boosted & SNIPS & $-3.5 \pm 4.7$ & $-0.3 \pm 4.7$ & $+2.8 \pm 5.8$ \\
boosted & DR & $-7.3 \pm 4.8$ & $-0.4 \pm 4.7$ & $+1.7 \pm 6.1$ \\
boosted & pilot-DR & $-7.5 \pm 13.1$ & $-10.1 \pm 13.3$ & $+13.9 \pm 14.4$ \\
\bottomrule
\end{tabular}
\end{adjustbox}
\end{table}

\begin{table}[!htbp]
\centering
\caption{Estimator score errors in condition $B_{\mathrm{on}}$.}
\label{tab:campaignfullerrorsB_on}
\small
\begin{adjustbox}{max width=\linewidth}
\begin{tabular}{llrrr}
\toprule
regime & estimator & $\pi_{\mathrm{visit}}$ & $\pi_{\mathrm{sms}}$ & $\pi_{\mathrm{stage}}$ \\
\multicolumn{2}{c}{} & \multicolumn{3}{c}{Mean error (US\$/decision)} \\
\midrule
tabular & DM & $-56.6 \pm 3.6$ & $+4.9 \pm 3.7$ & $-10.8 \pm 6.1$ \\
tabular & DM, context-blind & $-56.6 \pm 3.3$ & $+11.8 \pm 4.1$ & $-9.4 \pm 5.2$ \\
tabular & IS & $-56.7 \pm 3.6$ & $+4.7 \pm 3.7$ & $-10.9 \pm 6.1$ \\
tabular & SNIPS & $-56.6 \pm 3.6$ & $+4.9 \pm 3.7$ & $-10.8 \pm 6.1$ \\
tabular & DR & $-56.7 \pm 3.6$ & $+5.1 \pm 3.8$ & $-10.3 \pm 6.1$ \\
tabular & DM\textsuperscript{+}-IS & $-56.5 \pm 3.6$ & $+5.0 \pm 3.7$ & $-10.7 \pm 6.1$ \\
tabular & pilot-IS & $-8.6 \pm 11.5$ & $-4.6 \pm 18.5$ & $+15.3 \pm 14.6$ \\
tabular & pilot-DR & $-8.6 \pm 11.0$ & $-3.3 \pm 17.5$ & $+13.0 \pm 13.4$ \\
boosted & DM & $-50.3 \pm 2.0$ & $+3.9 \pm 1.8$ & $-9.3 \pm 2.2$ \\
boosted & IS & $-42.9 \pm 3.9$ & $+17.5 \pm 6.8$ & $-3.8 \pm 6.7$ \\
boosted & SNIPS & $-52.3 \pm 3.8$ & $+6.8 \pm 6.6$ & $-5.8 \pm 6.6$ \\
boosted & DR & $-53.8 \pm 4.0$ & $+6.7 \pm 6.5$ & $-5.9 \pm 7.0$ \\
boosted & pilot-DR & $-8.8 \pm 11.2$ & $-3.2 \pm 18.1$ & $+12.1 \pm 14.1$ \\
\bottomrule
\end{tabular}
\end{adjustbox}
\end{table}

Table~\ref{tab:campaignrmse} reports root-mean-square error with a
bootstrap-over-seeds SE for the representative estimators of both
regimes and the pilots. Under coupling the pilots have no uniformly lower
RMSE than the log-based estimators for the visit policy, because their
seed-level dispersion is of the same order as the log-based error, and
their RMSE is larger for the SMS and stage policies.

\begin{table}[!htbp]
\centering
\caption{Root-mean-square error by condition and target.}
\label{tab:campaignrmse}
\small
\begin{adjustbox}{max width=\linewidth}
\begin{tabular}{llrrrrr}
\toprule
condition & target & tabular DR & boosted DM & boosted DR & pilot-IS & pilot-DR \\
\multicolumn{2}{c}{} & \multicolumn{5}{c}{RMSE (US\$/decision)} \\
\midrule
$A_{\mathrm{off}}$ & $\pi_{\mathrm{visit}}$ & $21.1 \pm 2.8$ & $9.2 \pm 1.9$ & $22.5 \pm 3.3$ & $60.3 \pm 10.9$ & $57.4 \pm 11.3$ \\
$A_{\mathrm{off}}$ & $\pi_{\mathrm{sms}}$ & $14.0 \pm 1.7$ & $6.0 \pm 0.7$ & $25.0 \pm 4.3$ & $63.7 \pm 7.2$ & $59.5 \pm 6.6$ \\
$A_{\mathrm{off}}$ & $\pi_{\mathrm{stage}}$ & $26.6 \pm 4.0$ & $11.4 \pm 1.6$ & $26.8 \pm 3.8$ & $59.4 \pm 9.1$ & $54.2 \pm 8.0$ \\
$A_{\mathrm{on}}$ & $\pi_{\mathrm{visit}}$ & $59.8 \pm 3.6$ & $53.6 \pm 1.9$ & $57.9 \pm 3.7$ & $61.6 \pm 11.2$ & $59.5 \pm 11.3$ \\
$A_{\mathrm{on}}$ & $\pi_{\mathrm{sms}}$ & $17.6 \pm 2.2$ & $8.7 \pm 1.3$ & $29.1 \pm 6.6$ & $79.1 \pm 9.7$ & $75.4 \pm 9.0$ \\
$A_{\mathrm{on}}$ & $\pi_{\mathrm{stage}}$ & $32.0 \pm 4.0$ & $13.4 \pm 2.6$ & $33.0 \pm 3.4$ & $67.6 \pm 8.8$ & $63.4 \pm 9.0$ \\
$B_{\mathrm{off}}$ & $\pi_{\mathrm{visit}}$ & $20.7 \pm 2.5$ & $9.2 \pm 1.7$ & $22.3 \pm 2.2$ & $59.5 \pm 10.8$ & $56.3 \pm 11.0$ \\
$B_{\mathrm{off}}$ & $\pi_{\mathrm{sms}}$ & $13.3 \pm 1.7$ & $5.2 \pm 0.4$ & $20.4 \pm 2.9$ & $60.3 \pm 5.8$ & $55.7 \pm 5.5$ \\
$B_{\mathrm{off}}$ & $\pi_{\mathrm{stage}}$ & $24.0 \pm 3.6$ & $15.3 \pm 1.4$ & $26.5 \pm 2.0$ & $68.5 \pm 8.6$ & $62.1 \pm 7.6$ \\
$B_{\mathrm{on}}$ & $\pi_{\mathrm{visit}}$ & $58.9 \pm 3.4$ & $51.0 \pm 1.9$ & $56.6 \pm 3.6$ & $50.9 \pm 8.3$ & $48.7 \pm 8.2$ \\
$B_{\mathrm{on}}$ & $\pi_{\mathrm{sms}}$ & $17.4 \pm 2.7$ & $8.7 \pm 1.4$ & $29.1 \pm 7.3$ & $80.6 \pm 8.5$ & $76.2 \pm 7.8$ \\
$B_{\mathrm{on}}$ & $\pi_{\mathrm{stage}}$ & $28.5 \pm 4.4$ & $13.3 \pm 2.1$ & $31.1 \pm 3.8$ & $65.3 \pm 7.5$ & $59.9 \pm 7.6$ \\
\bottomrule
\end{tabular}
\end{adjustbox}
\end{table}

\FloatBarrier

\subsection{Numerical approximations and support}
Targets use discount-bin centers, whereas matching logged offers vary within the bin. A development comparison of the center dose against a uniform dose within the bin changes SMS and visit values by approximately 0.03 and 0.08~US\$ per decision.

A complementary first-order calculation holds recovery probabilities fixed and includes the reward factor $(1-d)$, weighted by context. The resulting logged-dose offsets are approximately $+0.45$~US\$ for the visit target, $-0.44$~US\$ for SMS and $-1.97$~US\$ for the stage-dependent target. Because this offset calculation excludes the recovery response to dose, it is not a complete bound on discretization error.

The action-support check finds target cells in every occupied context except a small set of extreme-delinquency contexts containing at most 0.07\% of the cohort. Randomized target matches range from 1,279 to 1,425 per run. Boosted estimated propensities are clipped at 0.001, below the design floor $\varepsilon/44\simeq0.0018$. Table~\ref{tab:campaignsupport} reports, per condition and target, the logged rows matching the target, the randomized matches, the contexts without support and their cohort share, the matched rows whose estimated propensity fell below the clip, the minimum estimated propensity and the mean boosted IS weight, whose excess over one is the Jensen inflation behind the positive boosted IS control error. Neither regime receives borrower-level latent factors.

\begin{table}[!htbp]
\centering
\caption{Support, clipping and importance weights in the paired study.}
\label{tab:campaignsupport}
\small
\begin{adjustbox}{max width=\linewidth}
\begin{tabular}{llrrrrrrr}
\toprule
condition & target & matched rows & randomized matches & unsupported contexts & cohort share (\%) & clipped matches & minimum propensity & mean IS weight \\
\midrule
$A_{\mathrm{off}}$ & $\pi_{\mathrm{visit}}$ & 9,744--10,111 & 1,279--1,405 & 7 & 0.07 & 5 & 0.00035 & 1.06 \\
$A_{\mathrm{off}}$ & $\pi_{\mathrm{sms}}$ & 30,566--31,087 & 1,309--1,425 & 7 & 0.07 & 31 & 0.00033 & 1.03 \\
$A_{\mathrm{off}}$ & $\pi_{\mathrm{stage}}$ & 12,475--12,887 & 1,318--1,418 & 7 & 0.07 & 6 & 0.00046 & 1.02 \\
$A_{\mathrm{on}}$ & $\pi_{\mathrm{visit}}$ & 12,436--12,814 & 1,279--1,405 & 7 & 0.07 & 4 & 0.00039 & 1.04 \\
$A_{\mathrm{on}}$ & $\pi_{\mathrm{sms}}$ & 28,730--29,356 & 1,309--1,425 & 7 & 0.07 & 35 & 0.00024 & 1.03 \\
$A_{\mathrm{on}}$ & $\pi_{\mathrm{stage}}$ & 13,253--13,716 & 1,318--1,418 & 7 & 0.07 & 4 & 0.00042 & 1.01 \\
$B_{\mathrm{off}}$ & $\pi_{\mathrm{visit}}$ & 9,744--10,111 & 1,279--1,405 & 7 & 0.07 & 5 & 0.00035 & 1.06 \\
$B_{\mathrm{off}}$ & $\pi_{\mathrm{sms}}$ & 30,566--31,087 & 1,309--1,425 & 7 & 0.07 & 31 & 0.00033 & 1.03 \\
$B_{\mathrm{off}}$ & $\pi_{\mathrm{stage}}$ & 12,475--12,887 & 1,318--1,418 & 7 & 0.07 & 6 & 0.00046 & 1.02 \\
$B_{\mathrm{on}}$ & $\pi_{\mathrm{visit}}$ & 12,436--12,814 & 1,279--1,405 & 7 & 0.07 & 4 & 0.00039 & 1.04 \\
$B_{\mathrm{on}}$ & $\pi_{\mathrm{sms}}$ & 28,730--29,356 & 1,309--1,425 & 7 & 0.07 & 35 & 0.00024 & 1.03 \\
$B_{\mathrm{on}}$ & $\pi_{\mathrm{stage}}$ & 13,253--13,716 & 1,318--1,418 & 7 & 0.07 & 4 & 0.00042 & 1.01 \\
\bottomrule
\end{tabular}
\end{adjustbox}
\end{table}

\subsection{Results under simulated cash}\label{sec:cashextra}

Tables~\ref{tab:campaignvisit_cash}, \ref{tab:campaignsms_cash} and
\ref{tab:campaignstage_cash} give mean cash errors and seed-level SEs in
US\$ per decision for the estimators of Table~\ref{tab:campaignvisit},
against independent cash references from the same simulations as the
score. Table~\ref{tab:campaignintervals_cash} repeats the interval
diagnostic under cash. Boosted IS keeps its positive control offset, and
the coupled visit error of the log-based estimators is close to its score
counterpart.

\begin{table}[!htbp]
\centering
\caption{$\pi_{\mathrm{visit}}$ errors by condition under cash.}
\label{tab:campaignvisit_cash}
\footnotesize
\begin{adjustbox}{max width=\linewidth}
\begin{tabular}{lrrrr}
\toprule
estimator & $A_{\mathrm{off}}$ & $A_{\mathrm{on}}$ & $B_{\mathrm{off}}$ & $B_{\mathrm{on}}$ \\
\multicolumn{1}{c}{} & \multicolumn{4}{c}{Mean error (US\$/decision)} \\
\midrule
tabular DR & $-1.1 \pm 4.4$ & $-49.6 \pm 3.5$ & $-4.0 \pm 4.2$ & $-49.1 \pm 3.2$ \\
boosted DM & $-4.9 \pm 1.6$ & $-46.5 \pm 1.6$ & $-4.6 \pm 1.6$ & $-44.5 \pm 1.7$ \\
boosted IS & $+12.4 \pm 5.0$ & $-38.0 \pm 3.7$ & $+9.0 \pm 4.8$ & $-37.5 \pm 3.5$ \\
boosted SNIPS & $-0.1 \pm 4.7$ & $-45.9 \pm 3.6$ & $-2.9 \pm 4.6$ & $-45.2 \pm 3.4$ \\
boosted DR & $-2.9 \pm 4.9$ & $-47.3 \pm 3.8$ & $-5.7 \pm 4.8$ & $-46.4 \pm 3.7$ \\
pilot-IS & $-2.6 \pm 12.2$ & $-2.5 \pm 11.6$ & $-5.1 \pm 11.7$ & $-6.9 \pm 9.4$ \\
tabular pilot-DR & $-2.0 \pm 11.8$ & $-2.5 \pm 11.3$ & $-4.5 \pm 11.2$ & $-6.9 \pm 9.0$ \\
\bottomrule
\end{tabular}
\end{adjustbox}
\end{table}

\begin{table}[!htbp]
\centering
\caption{$\pi_{\mathrm{sms}}$ errors by condition under cash.}
\label{tab:campaignsms_cash}
\footnotesize
\begin{adjustbox}{max width=\linewidth}
\begin{tabular}{lrrrr}
\toprule
estimator & $A_{\mathrm{off}}$ & $A_{\mathrm{on}}$ & $B_{\mathrm{off}}$ & $B_{\mathrm{on}}$ \\
\multicolumn{1}{c}{} & \multicolumn{4}{c}{Mean error (US\$/decision)} \\
\midrule
tabular DR & $+3.9 \pm 2.1$ & $+5.1 \pm 2.7$ & $+2.9 \pm 2.4$ & $+5.2 \pm 2.6$ \\
boosted DM & $+2.0 \pm 1.0$ & $+4.3 \pm 1.2$ & $+1.2 \pm 0.8$ & $+4.7 \pm 1.1$ \\
boosted IS & $+8.9 \pm 3.5$ & $+12.4 \pm 4.6$ & $+7.8 \pm 2.9$ & $+13.0 \pm 4.6$ \\
boosted SNIPS & $+2.0 \pm 3.4$ & $+5.0 \pm 4.5$ & $+0.9 \pm 2.9$ & $+5.5 \pm 4.5$ \\
boosted DR & $+1.9 \pm 3.3$ & $+5.0 \pm 4.4$ & $+1.0 \pm 2.7$ & $+5.4 \pm 4.4$ \\
pilot-IS & $-1.8 \pm 10.5$ & $+0.8 \pm 13.0$ & $-4.0 \pm 10.4$ & $+0.2 \pm 13.5$ \\
tabular pilot-DR & $-0.9 \pm 9.8$ & $+1.7 \pm 12.5$ & $-3.4 \pm 9.7$ & $+1.1 \pm 12.8$ \\
\bottomrule
\end{tabular}
\end{adjustbox}
\end{table}

\begin{table}[!htbp]
\centering
\caption{$\pi_{\mathrm{stage}}$ errors by condition under cash.}
\label{tab:campaignstage_cash}
\footnotesize
\begin{adjustbox}{max width=\linewidth}
\begin{tabular}{lrrrr}
\toprule
estimator & $A_{\mathrm{off}}$ & $A_{\mathrm{on}}$ & $B_{\mathrm{off}}$ & $B_{\mathrm{on}}$ \\
\multicolumn{1}{c}{} & \multicolumn{4}{c}{Mean error (US\$/decision)} \\
\midrule
tabular DR & $+2.9 \pm 4.1$ & $-8.1 \pm 4.2$ & $+1.4 \pm 3.6$ & $-7.6 \pm 3.7$ \\
boosted DM & $-0.4 \pm 1.1$ & $-1.1 \pm 1.1$ & $-4.9 \pm 1.3$ & $-3.7 \pm 1.0$ \\
boosted IS & $+7.0 \pm 3.8$ & $-6.3 \pm 4.5$ & $+6.1 \pm 3.4$ & $-4.6 \pm 4.1$ \\
boosted SNIPS & $+2.8 \pm 3.6$ & $-7.6 \pm 4.4$ & $+1.9 \pm 3.2$ & $-5.9 \pm 4.1$ \\
boosted DR & $+1.2 \pm 3.8$ & $-8.1 \pm 4.7$ & $+0.6 \pm 3.3$ & $-6.6 \pm 4.3$ \\
pilot-IS & $-1.5 \pm 8.8$ & $-1.2 \pm 9.9$ & $+4.5 \pm 9.6$ & $+2.6 \pm 9.4$ \\
tabular pilot-DR & $-2.8 \pm 8.1$ & $-2.7 \pm 9.3$ & $+3.4 \pm 8.8$ & $+1.4 \pm 8.6$ \\
\bottomrule
\end{tabular}
\end{adjustbox}
\end{table}

\begin{table}[!htbp]
\centering
\caption{Interval containment and width in the paired study under cash.}
\label{tab:campaignintervals_cash}
\footnotesize
\begin{adjustbox}{max width=\linewidth}
\begin{tabular}{llrrrr}
\toprule
condition & target & DM (fixed model) & DM refit & DR & pilot-IS \\
\multicolumn{2}{c}{} & \multicolumn{4}{c}{Contained / runs (mean width in US\$/decision)} \\
\midrule
$A_{\mathrm{off}}$ & $\pi_{\mathrm{visit}}$ & 1/20 (0.8) & 18/20 (61.7) & 18/20 (61.9) & 18/20 (161.6) \\
$A_{\mathrm{off}}$ & $\pi_{\mathrm{sms}}$ & 1/20 (0.8) & 19/20 (46.4) & 19/20 (47.2) & 17/20 (157.9) \\
$A_{\mathrm{off}}$ & $\pi_{\mathrm{stage}}$ & 0/20 (0.8) & 18/20 (60.8) & 18/20 (60.4) & 17/20 (150.2) \\
$A_{\mathrm{on}}$ & $\pi_{\mathrm{visit}}$ & 0/20 (0.7) & 2/20 (50.6) & 2/20 (50.5) & 18/20 (160.9) \\
$A_{\mathrm{on}}$ & $\pi_{\mathrm{sms}}$ & 0/20 (0.8) & 20/20 (47.5) & 20/20 (47.7) & 16/20 (161.1) \\
$A_{\mathrm{on}}$ & $\pi_{\mathrm{stage}}$ & 0/20 (0.8) & 18/20 (60.5) & 18/20 (60.5) & 18/20 (150.7) \\
$B_{\mathrm{off}}$ & $\pi_{\mathrm{visit}}$ & 0/20 (0.8) & 18/20 (60.8) & 17/20 (60.9) & 18/20 (154.9) \\
$B_{\mathrm{off}}$ & $\pi_{\mathrm{sms}}$ & 0/20 (0.8) & 19/20 (45.1) & 19/20 (45.7) & 18/20 (155.5) \\
$B_{\mathrm{off}}$ & $\pi_{\mathrm{stage}}$ & 1/20 (0.8) & 20/20 (61.0) & 20/20 (61.1) & 18/20 (153.5) \\
$B_{\mathrm{on}}$ & $\pi_{\mathrm{visit}}$ & 0/20 (0.6) & 2/20 (50.3) & 2/20 (50.2) & 18/20 (154.4) \\
$B_{\mathrm{on}}$ & $\pi_{\mathrm{sms}}$ & 0/20 (0.8) & 20/20 (46.3) & 20/20 (46.4) & 15/20 (160.7) \\
$B_{\mathrm{on}}$ & $\pi_{\mathrm{stage}}$ & 1/20 (0.8) & 19/20 (61.5) & 19/20 (61.9) & 17/20 (153.9) \\
\bottomrule
\end{tabular}
\end{adjustbox}
\end{table}

\FloatBarrier

\section{Score equivalence and auxiliary analyses}\label{sec:scoreequivalence}

The auxiliary checks examine reference precision, paired error contrasts
and score equivalence between simulator models.

\subsection{Reference precision and paired contrasts}\label{sec:referencechecks}

A post hoc comparison of reference-replicate spread against individual-loan
SE predictions gives variance ratios of 0.53--0.82 for values and
0.66--1.30 for paired gaps. Smaller-population checks found no stream or
variance-formula defect. In the grid's 20-seed cells, which use the same
reference design, the value ratios are 1.03--1.41, so the lower
paired-study value dispersion remains unexplained and is not a general
property of the references. In the paired study, reported value SEs
exceed the observed spread, without
independently establishing exact calibration.

For the same seed blocks, the paired on--off visit-error contrast for tabular
DR is
$-56.4\pm3.9$~US\$ in A and $-52.7\pm4.0$~US\$ in B. Boosted DR gives
$-51.4\pm4.6$ and $-46.5\pm4.7$~US\$. Covariates and nuisance fitting
change jointly with model class, preventing separate attribution.

\subsection{Simulator-score equivalence}

Model 2.0.0 permits partial payment and residual balances. Model 2.1.0
pays the full offer and closes the loan. Until first recovery, their configured streams and event probabilities
are identical. The score,
initial covariates, actions, annotations and references read nothing
after that event, and fitting/bootstrap seeds are unchanged. Equivalence
holds for the candidate policies of this study, which act at month zero
and read nothing after first recovery. Tools that consume draws after a
recovery are outside the argument.

To verify this score-equivalence argument, isolated comparisons cover the
first two batches of eight million-borrower
worlds, four conditions and two seeds, with exploration on/off, all roles
and candidates, both reference windows (9 and 12 months) and the
database-generation path
(Table~\ref{tab:scoreequivalence}). All 80 complete runs were then
regenerated under 2.1.0, and every estimation, reference and selection
field of their score panels equals the 2.0.0 output, so the reported
paired-study scores hold under the released model. Counts are
array/field/summary comparisons, not independent experiments.

\begin{table}[!htbp]
\centering
\caption{Score equivalence between simulator models.}
\label{tab:scoreequivalence}
\small
\begin{tabular}{lr}
\toprule
validation quantity & count \\
\midrule
comparison units & 24 \\
exact checks, all equal & 3968 \\
cross-model checks, all equal & 3904 \\
complete runs regenerated & 80 of 80 \\
equal result fields in those runs (timing and memory excluded) & 63,040 of 63,040 \\
\bottomrule
\end{tabular}
\end{table}

Analyses that read the full panel after a recovery must use the model
under which the panel was generated.

The exploration-score comparison determines its replicate count on
development seeds and evaluates shared confirmation seeds. Measured
CPU-hours for the auxiliary analyses, including development diagnostics
not reported here, are given in Table~\ref{tab:auxiliarycompute},
excluding the main studies and untimed diagnostics.

\begin{table}[!htbp]
\centering
\caption{Computation for score analyses and model verification.}
\label{tab:auxiliarycompute}
\small
\begin{tabular}{lr}
\toprule
analysis & CPU-hours \\
\midrule
exploration effect (Section~\ref{sec:explorationscore}) and an unreported accounting variant & 8.16 \\
partial-payment reward diagnostic (not reported) & 1.01 \\
score-equivalence validation (Table~\ref{tab:scoreequivalence}) and a superseded cash diagnostic & 6.69 \\
\bottomrule
\end{tabular}
\end{table}

\end{document}